\documentclass[11pt]{article}
\usepackage[preprint]{acl}
\usepackage{longtable}
\usepackage{times}
\usepackage{latexsym}
\usepackage{booktabs}
\usepackage{hyperref}
\usepackage{multirow}
\usepackage{float}
\usepackage{tcolorbox}
\usepackage[T1]{fontenc}
\usepackage[utf8]{inputenc}

\usepackage{microtype}
\usepackage{xcolor}
\usepackage{amsmath}
\usepackage{inconsolata}

\usepackage{graphicx}
\usepackage{soul}
\usepackage[normalem]{ulem}
\tcbuselibrary{skins, breakable}
\usepackage{enumitem} % For compact bullet lists inside review boxes

\definecolor{boxorange}{HTML}{FFF4E0}
\definecolor{boxpurple}{HTML}{F3E8FF}
\definecolor{boxgreen}{HTML}{E8F5E9}
\definecolor{boxblue}{HTML}{E3F2FD}

\definecolor{titleorange}{HTML}{D35400}
\definecolor{titlepurple}{HTML}{7D3C98}
\definecolor{titlegreen}{HTML}{1E8449}
\definecolor{titleblue}{HTML}{21618C}

\tcbset{
    reviewcard/.style={
        enhanced,
        breakable,                % Allows a review to split across columns/pages if long
        boxrule=0pt,
        frame hidden,
        arc=2mm,
        boxsep=1mm,
        left=2.5mm, right=2.5mm, top=2.5mm, bottom=2.5mm,
        fontupper=\scriptsize\sffamily,
        before skip=3mm,
        after skip=3mm
    }
}

\newcommand{\modelbadge}[2]{%
    \vspace*{-1mm}
    \begin{flushright}
        \textcolor{#1}{\textbf{\textit{#2}}}
    \end{flushright}
    \vspace*{-1.5mm}
}
\title{One Reference or Many? Rethinking Multi-Reference Training for Intermediate-Entropy Text Generation}

\author{
\textbf{Maitreya Prafulla Chitale}$^{1}$\thanks{Equal contribution.} \quad 
\textbf{Ketaki Mangesh Shetye}$^{1}$\footnotemark[1] \quad \textbf{Yash More}$^{1}$ \\
\textbf{Harshit Gupta}$^{2}$ \quad 
\textbf{Manav Chaudhary}$^{3}$ \quad 
\textbf{Manish Shrivastava}$^{1}$ \quad
\textbf{Vasudeva Varma}$^{1}$ \\
$^1$IIIT Hyderabad \quad
$^2$Microsoft, India \quad
$^3$Sentisum \\
\texttt{\{maitreya.chitale, ketaki.shetye, yash.more\}@research.iiit.ac.in} \\
\texttt{harshit.g@alumni.iiit.ac.in} \\
\texttt{manav@sentisum.com} \\
\texttt{\{m.shrivastava, vv\}@iiit.ac.in}
}

\begin{document}
\maketitle
\begin{abstract}
Natural language generation (NLG) ranges from highly constrained tasks, such as machine translation, to open-ended tasks like dialogue generation. Between them lie peer-review generation and image captioning, where one input admits several valid, overlapping outputs. We measure this intermediary region with Von Neumann entropy ($\hat{H}\approx0.6$ vs.\ 0.15 translation, 0.92 dialogue). Single-reference training discards valid perspectives, while multi-reference training introduces high-variance, unaligned targets. To tackle this, we propose \emph{target-side consolidation}, that is, merging all references for an input into a single target which preserves every distinct information unit. To this end, we introduce \textbf{MERC-67K}, pairing 36{,}537 papers and 31{,}014 images with their original and consolidated references. In human evaluation, models trained using consolidated reviews obtain an average rank of 1.70 (out of 4) for coverage against 2.51--2.98 for baselines, and win 63.5--74.0\% of pairwise comparisons. Models trained using consolidated captions obtain an average rank of 1.87 for overall caption quality (accuracy and naturalness) against 2.64--2.82 for baselines and obtain the highest reference-free LongCLIPScore (0.839 vs best baseline's 0.816). For reviews, multi-reference training often under-performs zero-shot inference, and consolidation's gains over multi-reference training hold across Llama3.1, Mistralv0.3, and Qwen2.5. Consolidation uses $\sim$4$\times$ (reviews) and 5$\times$ (captions) fewer individual references during optimization. Dataset and code will be released upon acceptance.

\end{abstract}

\section{Introduction}
% Drawing on to the definitions of \citet{10.5555/3241691.3241693, 10.1145/3554727, 10.1017/S1351324997001502}, Natural Language Generation can be defined as: \textit{The subfield of artificial intelligence and computational linguistics that is concerned with the construction of computer systems that can produce understandable texts in English or other human languages from some underlying representation of information, which may be either non-linguistic or textual.}
\begin{figure*}[htbp]
    \centering
    \includegraphics[width=\linewidth]{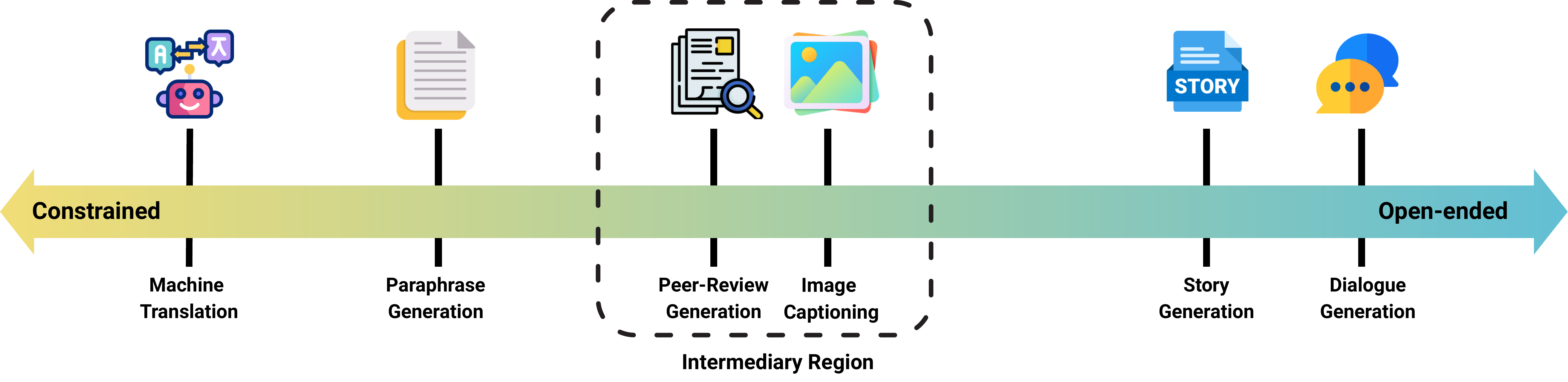}
    \caption{The NLG spectrum, classifying tasks by entropy from constrained to open-ended. The dashed box marks the intermediary region, which holds structured yet subjective tasks such as peer-review generation and image captioning.}
    \label{fig:spectrum}
\end{figure*}
Natural language generation (NLG) tasks span an entropy spectrum \citep{https://doi.org/10.1002/j.1538-7305.1948.tb01338.x, ijcai2022p818}. Low-entropy tasks like machine translation map inputs to narrow targets, while high-entropy open-ended tasks like dialogue generation permit highly diverse valid responses.

% Between these extremes lies an intermediary region of structured yet subjective tasks, such as Code Review, Question-Answer, and Reer-Review generation (Figure~\ref{fig:spectrum}). While strictly grounded in specific inputs, these tasks permit significant interpretive freedom. For instance, \citet{zheng-etal-2018-multi} show that shifting to multi-reference training doubles BLEU gains in Image Captioning compared to MT, while \citet{Vakulenko2023UniformTA} and \citet{zhao-kawahara-2021-multi} demonstrate the fundamental insufficiency of single-reference training for open-ended tasks like QA and Dialogue Generation. 

% Between these extremes lie structured yet subjective tasks like automated peer-review generation and image captioning (Figure~\ref{fig:spectrum}), which require specific input grounding while allowing significant interpretive freedom. Prior work shows multi-reference training outperforms single-reference baselines for tasks such as machine translation and image captioning while also highlighting that high reference variance within a single training batch can destabilize model learning for image captioning \citep{zheng-etal-2018-multi}. At the other end of the spectrum, relying solely on single-reference training remains fundamentally insufficient for open-ended generation \citep{zhao-kawahara-2021-multi}.

Between these extremes lie structured yet subjective tasks like automated peer-review generation and image captioning (Figure~\ref{fig:spectrum}), which require specific input grounding while allowing significant interpretive freedom. \citet{zheng-etal-2018-multi} show that multi-reference training outperforms single-reference baselines in machine translation and image captioning, but that high reference variance within a training batch can destabilize learning. At the open-ended extreme, \citet{zhao-kawahara-2021-multi} find that single-reference training is insufficient for dialogue response generation.

Despite these findings, training strategies for the intermediary domain remain unexplored. We argue that neither single- nor multi-reference approaches are optimal here. Valid references for a single input are semantically diverse (Section~\ref{sec:positioning}). Thus, a single reference discards perspectives, while unaligned multi-reference targets can introduce high variance within training batches \citep{zheng-etal-2018-multi}. To address this, we propose target-side consolidation, that is, merging all references for an input into one target which preserves every distinct information unit.
We investigate this paradigm through peer-review generation and image captioning, isolating the impact of consolidated targets in the intermediary NLG spectrum by comparing them directly against single- and multi-reference training on fixed base models. Our main contributions are:

\begin{itemize}
    % \item We introduce a large-scale corpus, \textbf{MERC-67K}, comprising over 36,000 papers across seven academic conferences, paired with both their original reviews and the consolidated reviews.
    % \item We show that consolidated-reference training outperforms single- and multi-reference training in human evaluation and annotators rank models trained on consolidated references best on average, at 1.70 (of 4) for review coverage vs.\ 2.51--2.98, and 1.87 for caption overall quality vs.\ 2.64--2.82.
    % \item We show that consolidated training outperforms single- and multi-reference baselines, achieving higher recall in automatic metrics and ranking first in human evaluations for review coverage (1.70 of 4) and caption quality (1.87).
    % \item We show that consolidated training outperforms single- and multi-reference baselines, ranking on average at 1.70 (of 4) for review coverage vs.\ 2.51--2.98, and 1.87 for caption overall quality vs.\ 2.64--2.82 human evaluations
     \item We show that consolidated-reference training yields more comprehensive reviews and higher-quality captions than single- and multi-reference training, with statistically significant automatic evaluation and moderate-to-substantial annotator agreement in human evaluation.
    \item We introduce \textbf{MERC-67K}, pairing $36,537$ papers with their original and consolidated reviews, and $31,014$ images with their five human captions and a consolidated caption.
    % \item We quantitatively place both tasks in the intermediary region of an entropy spectrum spanning six NLG tasks.
    % \item We benchmark state-of-the-art (SOTA) systems on this corpus, utilizing both extensive automatic metrics and human evaluation.
    % \item We demonstrate using automatic metrics and human evaluation that models trained on consolidated references significantly and consistently outperform those trained on non-consolidated references, providing empirical evidence for the superiority of reference consolidation in intermediary NLG tasks.
\end{itemize}

\section{Related Work}
\label{sec:related-works}

% Recent studies address the diversity of reviewer perspectives through multi-reference fine-tuning, such as filtering for high-confidence reviews to train Llama-OpenReviewer-8B \citep{idahl-ahmadi-2025-openreviewer}. To handle the high variance of human reviews, others focus on input-side conditioning: \citet{gao2024reviewer2} augment inputs with aspect-specific prompts to control coverage, while some distribute cognitive load across specialized agents \citep{darcy2024margmultiagentreviewgeneration, goyal2026scholarpeercontextawaremultiagentframework} or introduce hierarchical question-answering frameworks \citep{chang-etal-2025-treereview}.

% We argue that capturing the full distribution of valid critiques requires unifying diverse perspectives into a cohesive target. Recent frameworks train on such consolidated reviews, notably SEA \citep{yu-etal-2024-automated} and AutoRev \citep{chitale2026graphguidedpassageretrievalauthorcentric}. However, neither systematically benchmark the underlying training paradigms. We address this gap by providing empirical evidence on whether consolidated-signal training inherently outperforms traditional single- or multi-reference approaches in this intermediary regime.

% Prior work either conditions the input or builds consolidated data without isolating the effect of the training target, and this paper aims at investigating the same.
\subsection{Multi-Reference Training and Evaluation}
Training strategies leveraging multiple references have been studied across the entropy spectrum. In machine translation, relying on a single reference leaves most valid translations unobserved. To mitigate this, \citet{khayrallah-etal-2020-simulated} sample paraphrases of the target as extra references, and \citet{pmlr-v97-shen19c} use mixture models over valid translations. \citet{zheng-etal-2018-multi} compare ways of using multiple references in translation and captioning, finding gains over single reference training but noting that for captioning high reference variance within a batch may harm learning. For dialogue, single-reference training is insufficient \citep{zhao-kawahara-2021-multi} and multiple human references are needed for evaluation \citep{gupta-etal-2019-investigating}. However, these studies treat references as independent or augmented targets, not as one unified signal.

\subsection{Peer-Review Generation}
Recent work addresses reviewer diversity by fine-tuning on filtered human reviews, such as high-confidence reviews for OpenReviewer \citep{idahl-ahmadi-2025-openreviewer}. Others condition the input: \citet{gao2024reviewer2} add aspect-specific prompts to control coverage, while some distribute the task across specialized agents \citep{darcy2024margmultiagentreviewgeneration, goyal2026scholarpeercontextawaremultiagentframework} or introduce hierarchical question-answering frameworks \citep{chang-etal-2025-treereview}. Thus, these approaches modify either which reviews are selected or how the paper is processed.

\subsection{Image Captioning}
Multiple human captions can be equally valid for one image, yet models optimized against them produce low-diversity outputs \citep{dcfd163a97d7413e92b83a92cf723b09}. Recent work instead improves the training targets: CapsFusion \citep{Yu_2024_CVPR} uses an LLM to consolidate web-crawled and synthetic captions into one training caption, and ShareGPT4V \citep{chen2024sharegpt4v} trains large multimodal models on long, detailed captions. In both cases, the focus is on enriching or rewriting the captions.

\subsection{Target-Side Consolidation}
Consolidation unifies diverse perspectives into one cohesive target. Recent frameworks, notably SEA \citep{yu-etal-2024-automated} and AutoRev \citep{chitale2026graphguidedpassageretrievalauthorcentric}, train on consolidated reviews. A similar idea appears in opinion summarization, where \citet{brazinskas-etal-2020-unsupervised} generate summaries reflecting the consensus opinions of many reviews of the same product.

\subsection{Learning from Annotator Disagreement}
Whether to preserve or aggregate diverse human labels is also debated in annotation. \citet{plank-2022-problem} argues that label variation is a signal to be modeled rather than noise, and \citet{davani-etal-2022-dealing} show that modeling individual annotators can beat training on a majority-vote label. Consolidation differs from majority voting: instead of selecting one viewpoint, it merges complementary points from all references into one target. Whether this yields a superior training signal is the core question of this work.

\subsection{Research Gap}
Prior work leaves three gaps for the intermediary region. First, multi-reference training has mainly been studied where outputs are tightly constrained (machine translation) or fully open-ended (dialogue), with references treated as independent or augmented targets. But how it behaves when one input admits multiple overlapping, complementary outputs (intermediary region) is largely unexamined. Second, the specific benefits of existing consolidated targets (SEA, AutoRev, CapsFusion, ShareGPT4V) cannot be isolated as they are coupled with larger architectures. To our knowledge, no prior work compares consolidated, single- and multi-reference training using identical inputs and base model. Third, although the annotator-disagreement literature resolves diverse human perspectives by selecting or distributing classification labels, intermediary NLG tasks require synthesizing these overlapping perspectives into cohesive free-form text. We address these gaps by varying the supervision on a fixed base model and data, for both the tasks, using our \textbf{MERC-67K} dataset.

\section{Positioning Intermediary Tasks}
\label{sec:positioning}
This section quantitatively maps the position of ``intermediary tasks'' in the NLG entropy spectrum (Figure~\ref{fig:spectrum}). We aim to measure the inherent semantic diversity of these tasks by analyzing the variance among valid reference outputs conditioned on a single source input. Towards this, we use the von Neumann entropy, the quantity underlying the Vendi Score introduced by \cite{friedman2023vendi}. As our objective is to evaluate local semantic variance among multiple references, we calculate the entropy strictly at the instance level rather than aggregating a corpus-wide Vendi Score as originally done by \citet{friedman2023vendi}. To quantify the entropy across this spectrum, we use several established multi-reference datasets for six diverse generation tasks (Table~\ref{tab:entropy_tasks}). Further details for the chosen datasets are provided in Appendix~\ref{app:entropy-dataset}.

% Specifically, we evaluate Machine Translation (English--German) on \citet{ea-mt-benchmark, conia-etal-2024-towards}, Paraphrase Generation on \citet{10.1007/978-3-030-77385-4_36}, Peer Review Generation on our MERC-67K dataset, Image Captioning on \citet{7410660}, Story Generation on \citet{fan-etal-2018-hierarchical}, and Dialogue Generation on \citet{sai-etal-2020-improving}.
For each instance, we extract sentence embeddings for all $N$ reference texts using \textit{BAAI/bge-m3} \citep{chen-etal-2024-m3}. We compute their pairwise cosine-similarity matrix and normalize it by $N$ to yield a density matrix ($\rho$), with a trace of 1. The raw von Neumann entropy, $S$, is then calculated using the eigenvalues $\lambda_i$ of $\rho$:

$$S(\rho) = -\sum_i \lambda_i \log_2 \lambda_i$$

To fairly compare tasks, we introduce empirically normalized entropy ($\hat{H}$). We normalize $S$ against $S_{\text{random}}$, the expected raw entropy of $N$ texts drawn uniformly at random from a task's global reference pool. This normalization is necessary because dense text embeddings occupy a non-uniform space where mutual orthogonality is practically unattainable, even for unrelated instances \citep{ethayarajh-2019-contextual,li-etal-2020-sentence}. Thus, $S_{\text{random}}$ establishes an upper bound for the maximum diversity realistically achievable.

$$\hat{H} = \frac{S}{S_{\text{random}}}$$

Lower $\hat{H}$ values indicate concentrated, semantically similar references, while higher values indicate broad diversity. 

As shown in Table~\ref{tab:entropy_tasks}, constrained tasks like translation and paraphrase generation exhibit low entropy, reflecting a narrow space of valid outputs. At the opposite extreme, open-ended tasks like dialogue and story generation approach maximum possible empirical randomness. These results indicate that both tasks exhibit moderate semantic variance, empirically placing them in the intermediary region.
\begin{table}[htbp]
\centering
\begin{tabular}{lc}
\toprule
\textbf{Task} & \begin{tabular}[c]{@{}c@{}}\textbf{Entropy}\\ \textbf{($\hat{H}$)}\end{tabular} \\
\midrule
Translation & 0.15 \\
Paraphrase Generation & 0.35 \\
Peer-Review Generation & 0.61 \\
Image Captioning & 0.62  \\
Story Generation & 0.90  \\
Dialogue Generation & 0.92  \\
\bottomrule
\end{tabular}
\caption{Average empirically normalized Von Neumann entropy ($\hat{H} = S / S_{\text{random}}$) across different NLG tasks. Higher values indicate more diverse references.}
\label{tab:entropy_tasks}
\end{table}

\section{The MERC-67K Dataset}

% \begin{table}[htbp]
% \centering
%     \begin{tabular}{lcc}
%     \toprule
%     \textbf{Conference} & \textbf{\# Papers} & \textbf{\# Reviews} \\ \midrule
%     COLM 2024           & 293            & 1127            \\
%     COLM 2025           & 405            & 1537            \\
%     NeurIPS 2023        & 3358           & 14989           \\
%     NeurIPS 2024        & 4149           & 16299           \\
%     ICLR 2024           & 5634           & 21795           \\
%     ICLR 2025           & 8268           & 33574           \\
%     ICLR 2026           & 14430          & 56209           \\ \midrule
%     \textbf{Total}      & \textbf{36537} & \textbf{145530} \\ \bottomrule
%     \end{tabular}
% \caption{Dataset statistics for the MERC-67K corpus, detailing the distribution of scraped papers and their corresponding reviews across seven major machine learning conferences.}
% \label{tab:pdf-stats-short}
% \end{table}

\begin{table}[htbp]
\centering
% \small
\begin{tabular}{lccc}
\toprule
\textbf{Venue} & \textbf{Train} & \textbf{Val} & \textbf{Test} \\
\midrule
\multicolumn{4}{l}{\textit{\textbf{Peer-Review Generation (Papers)}}} \\
\midrule
COLM 24 & 234 & 29 & 30 \\
COLM 25 & 324 & 40 & 41 \\
NeurIPS 23 & 2,686 & 336 & 336 \\
NeurIPS 24 & 3,319 & 415 & 415 \\
ICLR 24 & 4,507 & 564 & 563 \\
ICLR 25 & 6,610 & 827 & 831 \\
ICLR 26 & 11,544 & 1,443 & 1,443 \\
\midrule
\textbf{Subtotal} & \textbf{29,224} & \textbf{3,654} & \textbf{3,659} \\
\midrule
\multicolumn{4}{l}{\textit{\textbf{Image Captioning (Images)}}} \\
\midrule
Flickr30k & 29,000 & 1,014 & 1,000 \\
\midrule
% \textbf{Subtotal} & \textbf{29,000} & \textbf{1,014} & \textbf{1,000} \\
\midrule
\textbf{Total Instances} & \multicolumn{3}{c}{\textbf{67,551}} \\
\bottomrule
\end{tabular}
\caption{Papers and images per split in MERC-67K. Reviews use a venue-stratified 80-10-10 split and captions use the Karpathy split.}
\label{tab:combined_dataset_splits}
\end{table}

% \begin{table*}[htbp]
% \centering
%     \begin{tabular}{lccc}
%     \toprule
%     \textbf{Conference} & \textbf{Total Papers} & \textbf{Total Reviews} &\textbf{Avg Reviews/Paper}\\ \midrule
%     COLM 2024             & 293    & 1127 & 3.85            \\
%     COLM 2025             & 405 & 1537 & 3.80                \\
%     NeurIPS 2023          & 3358 & 14989 & 4.46               \\
%     NeurIPS 2024          & 4149 & 16299 & 3.93               \\
%     ICLR 2024             & 5634 & 21795 & 3.87               \\
%     ICLR 2025             & 8268 & 33574 & 4.06               \\
%     ICLR 2026             & 14430 & 56209 & 3.85              \\ \midrule
%     \textbf{Total}        & \textbf{36537}   & \textbf{145530} & \textbf{3.98}  \\ \bottomrule
%     \end{tabular}
% \caption{Summary of scraping statistics across target conferences.}
% \label{tab:pdf-stats-short}
% \end{table*}
% In this section, we introduce MERC-67K: \textbf{M}id-\textbf{E}ntropy \textbf{R}eference \textbf{C}onsolidation dataset, comprising 36,537 papers, paired with their original peer reviews from OpenReview, and their corresponding structured and consolidated review.

We introduce MERC-67K, the \textbf{M}id-\textbf{E}ntropy \textbf{R}eference \textbf{C}onsolidation dataset. This dataset comprises two subsets: $36,537$ academic papers paired with their original and consolidated peer reviews for the peer-review generation task, and $31,014$ images paired with their original human-written and consolidated captions for the image captioning task.     
\subsection{Data Collection \& Normalization}
% We collect PDF submissions and official peer reviews from OpenReview (ICLR, NeurIPS, and COLM; Table~\ref{tab:pdf-stats-short}), converting the PDFs into Multi Mark Down (MMD) following \citet{yu-etal-2024-automated}. 
% To address varying reviewing guidelines across venues, we normalize the extracted reviews into a uniform structure. Following \citet{chitale2026graphguidedpassageretrievalauthorcentric}, we exclude numerical scores to focus purely on qualitative textual critique, standardizing all reviews into four core sections: \textit{Summary}, \textit{Strengths}, \textit{Weaknesses}, and \textit{Questions}.
For the peer-review generation task, we scrape manuscripts and official reviews from OpenReview (ICLR, NeurIPS, COLM; Statistics in Table \ref{tab:combined_dataset_splits}) and convert PDFs to Multi Mark Down (MMD) following \citet{yu-etal-2024-automated}. To handle varying venue guidelines, we exclude numerical scores and standardize textual critiques into four core sections: Summary, Strengths, Weaknesses, and Questions \citep{chitale2026graphguidedpassageretrievalauthorcentric}. We partition the Peer-Review Generation subset of the dataset into a venue-stratified 80-10-10 train-validation-test split. For the image captioning task, we use the \textit{Flickr30k dataset} \citep{young-etal-2014-image}, comprising $31,014$ images originally paired with five independent human reference captions. We adopt the established Karpathy split \citep{Karpathy_2015_CVPR} to benchmark our consolidation-focused training approach against standard multi-reference training.

% \paragraph{Consolidation}
% To generate the consolidated reviews, we utilize the SEA-S model \citep{yu-etal-2024-automated}. The model synthesizes the diverse, multi-perspective reviewer critiques into a single, cohesive review that strictly mirrors our standardized four-section format.
% We utilize the SEA-S model \citep{yu-etal-2024-automated} to synthesize diverse reviewer critiques into a single, cohesive review that strictly follows our standardized four-section format.

% \paragraph{Formal Definition}
% Formally, each paper $p_i$ is associated with a set of $n$ original peer reviews scraped from OpenReview, denoted as $R_i = \{r_{i1},\,r_{i2},\,...,\,r_{in}\}$. These independent reviews are synthesized to form a single, unified consolidated review $c_i$. Consequently, our final dataset consists of 36,537 unique tuples of the form $(p_i,\, R_i,\, c_i)$. The final statistics can be found in Table \ref{tab:pdf-stats-short}.
% Each paper $p_i$ is associated with $n$ original peer reviews $R_i = \{r_{i1}, r_{i2}, \dots, r_{in}\}$, which are synthesized into a single consolidated review $c_i$. Our final dataset comprises 36,537 unique $(p_i, R_i, c_i)$ tuples (see Table~\ref{tab:pdf-stats-short}).

\subsection{Consolidation \& Formal Definition}
Given an input $x$ and its references $R=\{r_1,\dots,r_n\}$, \emph{consolidation} produces a single target $c$ that (i) covers every distinct information unit in $R$, (ii) states shared points exactly once, and (iii) introduces no content absent from $R$. Where references contradict one another, the most supported claim is retained. Unlike selecting one reference or majority voting, consolidation preserves all non-conflicting content, resulting in a target that is complete without being repetitive.
% states points shared across references once, and (iii) introduces no content absent from $R$. Where references conflict, the better-supported detail is retained. Unlike selecting one reference or majority voting, all non-conflicting content from $R$ is preserved, so the target is complete without being repetitive.

For peer-review generation, for each paper $p_{i}$, all the $n$ original reviews $R_{i}=\{r_{i1},r_{i2},...,r_{in}\}$ are consolidated into a single review $c_{i}$ using SEA-S \citep{yu-etal-2024-automated}, retaining only the \textit{Summary, Strengths, Weaknesses} and \textit{Questions} sections. In the original study, \textit{SEA-S} followed the required format in 100\% of cases (vs.\ 90.3\% for GPT-4), and expert annotators rated it above GPT-4, GPT-3.5 and Mistral-7B on all seven criteria: merging of similar points, objectivity, informativeness, low repetition, absence of contradiction, professional language, and overall quality. The final corpus contains $36,537$ $(p_{i},R_{i},c_{i})$ tuples.

For image captioning, each image $v_{j}$ is paired with $5$ human reference captions $C_{j}=\{c_{j1}, c_{j2}, \dots, c_{j5}\}$. We consolidate them into a single caption $\bar{c}_{j}$ using \textit{Llama-3.1-8B-Instruct}, prompted to preserve all non-conflicting details, add nothing absent from the captions, and favor the better-supported detail when captions disagree. The resulting corpus contains $31,014$ $(v_{j}, C_{j}, \bar{c}_{j})$ tuples. For more details, refer to Appendix~\ref{lst:caption-consolidate-prompt}.

\section{Methodology}
\label{sec:methodology}

We study peer-review generation and image captioning. For reviews, inputs are limited to the paper's main body as per conference guidelines, and our primary base model is \textit{Llama-3.1-8B-Instruct}~\footnote{https://ai.meta.com/blog/meta-llama-3-1/}, owing to its widespread use in peer-review generation. To test generalization across model families, we run ablations on a 3,000-paper training subset of MERC-67K using \textit{Qwen2.5-7B-Instruct}~\cite{qwen2025qwen25technicalreport} and \textit{Mistral-7B-Instruct-v0.3}~\cite{jiang2023mistral7b}, alongside a similarly trained \textit{Llama-3.1-8B-Instruct} model (Llama). For captioning, we use \textit{Molmo-7B-O}~\cite{11092508} (Molmo), whose pre-/post-training explicitly excludes Flickr30k. As the only such base model, it precludes cross-model ablations without data leakage. All models are capped at $8B$ parameters due to compute; details are in Appendix~\ref{sec:appendix-imp}.

We benchmark three paradigms: \textbf{Single-Reference} (one isolated target per input), \textbf{Multi-Reference} (multiple unaligned targets per input), and \textbf{Consolidation-Driven} (multiple references merged into a single reference). For a fair comparison, the multi-reference model trains on all N references, and we shuffle all data points to mitigate batch-level reference variance \citep{zheng-etal-2018-multi}. Because consolidation merges the points of several references, consolidated targets are longer than any single reference but shorter than the concatenation of multiple references. Longer outputs are an expected consequence of broader coverage, not an independent factor (Appendix~\ref{sec:appendix_length_analysis}).

% \textbf{Training Efficiency.} Consolidation uses fewer individual references during optimization, so it trains much faster than multi-reference training.

\textbf{Training Efficiency.} Consolidation uses fewer individual references during optimization ($\sim$4$\times$ for reviews, 5$\times$ for captions), reducing training cost relative to multi-reference training.

% \subsection{Peer Review Generation}
% \paragraph{Single-Reference:}Using the \textit{Llama-3.1-8B-Instruct} backbone, we benchmark \textit{Llama-Inference} (zero-shot) and \textit{Llama-HighConf} (fine-tuned solely on the highest-confidence original review per paper). 
% \paragraph{Multi-Reference:}  Using \textit{Reviewer2}~\citep{gao2024reviewer2}, \textit{OpenReviewer}~\citep{idahl-ahmadi-2025-openreviewer}, and \textit{Llama-MultiRef} (Llama backbone fine-tuned by treating each of the $n$ original reviews for a paper as an independent target).
% \paragraph{Consolidation-Driven:} Using \textit{SEA-E}~\citep{yu-etal-2024-automated}, \textit{AutoRev}~\citep{chitale2026graphguidedpassageretrievalauthorcentric}, and \textit{Llama-Consolidated} (Llama backbone fine-tuned on consolidated reviews).

% \subsection{Image Captioning}
% \paragraph{Single Reference:}Flickr30k has no natural single ``best'' caption analogous to high-confidence reviews, so we instead randomly sample a caption from the set of 5 human written captions and use that as the sole target. We call this variation \textit{Molmo-SingleRef}.
% \paragraph{Multi-Reference:}Using \textit{BLIP-2 OPT-6.7B}~\cite{pmlr-v202-li23q}, \textit{GIT-base}~\cite{wang2022git}, \textit{PaliGemma-3B}~\cite{beyer2024paligemmaversatile3bvlm}, and our \textit{Molmo-Multiref} (Molmo backbone fine-tuned by treating each of the $5$ original captions for an image as an independent target).
% \paragraph{Consolidation-Driven:}Using \textit{Molmo-Consolidated} (Molmo backbone fine-tuned on consolidated captions).
\subsection{Peer-Review Generation}
\paragraph{Single-Reference:} Within this group we use the \textit{Llama-3.1-8B-Instruct} base model for two single-reference systems: \textit{Llama-ZeroShot} (evaluated zero-shot) and \textit{Llama-HighConf} (fine-tuned solely on the highest-confidence original review for each paper; ties are broken by random choice).
\paragraph{Multi-Reference:} This group includes the existing baselines \textit{Reviewer2} \citep{gao2024reviewer2} and \textit{OpenReviewer} \citep{idahl-ahmadi-2025-openreviewer}, alongside our own \textit{Llama-MultiRef} (the Llama base model fine-tuned by treating each of the $n$ original reviews per paper as an independent target).
\paragraph{Consolidation-Driven:} In this group, we compare against \textit{SEA-E} \citep{yu-etal-2024-automated} and \textit{AutoRev} \citep{chitale2026graphguidedpassageretrievalauthorcentric}, which train on consolidated reviews, and introduce \textit{Llama-Consolidated} (our Llama base model fine-tuned directly on the consolidated target reviews).

\subsection{Image Captioning}
\paragraph{Single-Reference:} \textit{Molmo-SingleRef} is trained on one caption randomly sampled from the five human-written annotations, as the image captioning data lacks a natural ``best'' caption analogous to a high-confidence review.
% \paragraph{Multi-Reference:} To capture standard multi-target exposure, we evaluate established vision-language models including \textit{BLIP-2 OPT-6.7B}~\cite{pmlr-v202-li23q}, \textit{GIT-base}~\cite{wang2022git}, and \textit{PaliGemma-3B}~\cite{beyer2024paligemmaversatile3bvlm}. We contrast these with \textit{Molmo-Multiref}, our backbone fine-tuned by treating all five original captions for a given image as independent learning targets.
\paragraph{Multi-Reference:} This group includes the established vision-language models \textit{BLIP-2 OPT-6.7B}~\cite{pmlr-v202-li23q}, \textit{GIT-base}~\cite{wang2022git}, and \textit{PaliGemma-3B}~\cite{beyer2024paligemmaversatile3bvlm}. We also include our own \textit{Molmo-Multiref} (Molmo base model fine-tuned by treating all five original captions for a given image as independent learning targets).

\paragraph{Consolidation-Driven:} We test the efficacy of consolidated references using \textit{Molmo-Consolidated} (Molmo base model fine-tuned exclusively on the consolidated captions).
\paragraph{Long-Form Captioners:}
\label{sec:long-form-captioners}
We also include long-form captioners that generate detailed, multi-sentence descriptions: \textit{LLaVA-1.5-7B}~\cite{10655294}, \textit{ShareGPT4V-7B}~\cite{chen2024sharegpt4v}, and \textit{Molmo-Zeroshot}. These systems are evaluated off-the-shelf on Flickr30k and serve as a long-generation contrast to the short-form single-/multi-reference and consolidation-driven captioners above. No analogous long-form category exists for peer review, so this group appears only for captioning.

\subsection{Human Evaluation Design}
\label{main:human-eval-design}
As standard $n$-gram metrics and BERTScore often struggle with open-ended generation \citep{schluter-2017-limits, hanna-bojar-2021-fine}, we conduct a blinded-ranking study to evaluate both tasks. 

For peer-review generation, 30 expert ML researchers evaluate 50 papers uniformly sampled across venues. Three annotators per paper rank the outputs (1–4, allowing fractional ties) of four variants, \textit{Llama-ZeroShot}, \textit{Llama-HighConf}, \textit{Llama-MultiRef}, and \textit{Llama-Consolidated}, across four sections: \textit{Summary, Strengths, Weaknesses, and Questions}. This yields 600 ranked instances. Since SEA-S consolidation is rated highly by experts for merging overlapping reviewer points without losing any information \citep{yu-etal-2024-automated}, coverage of the original reviews is a meaningful criterion for comparing models (discarding valid perspectives is the failure mode we target). 

For image captioning, five annotators are provided with 50 randomly selected images and rank the captions generated by \textit{Molmo-ZeroShot}, \textit{Molmo-SingleRef}, \textit{Molmo-MultiRef}, and \textit{Molmo-Consolidated} on overall quality, i.e., how accurately and naturally each describes the image's most salient features without hallucinated details, using the same 1-4 scale.

Inter-annotator reliability is measured with Kendall's $W$~\citep{10.1214/aoms/1177732186}, which ranges from 0 (no agreement) to 1 (complete agreement). We obtain $W = 0.614$ for peer-review generation and $W = 0.487$ for image captioning. Following common practice \citep{Landis1977TheMO}, we describe W = 0.614 as substantial and W = 0.487 as moderate; these bands are heuristic. Complete annotation guidelines are provided in Appendix~\ref{app:human_eval}.
\begin{table*}[ht]
\centering
\resizebox{\textwidth}{!}{%
\begin{tabular}{ll ccc ccc ccc c}
\toprule
 & & \multicolumn{3}{c}{\textbf{F1}} & \multicolumn{3}{c}{\textbf{Precision}} & \multicolumn{3}{c}{\textbf{Recall}} & \\
\cmidrule(lr){3-5} \cmidrule(lr){6-8} \cmidrule(lr){9-11}
\textbf{Model} & \textbf{Ground Truth} & \textbf{R-1} & \textbf{R-2} & \textbf{R-L} & \textbf{R-1} & \textbf{R-2} & \textbf{R-L} & \textbf{R-1} & \textbf{R-2} & \textbf{R-L} & \textbf{BS-F1} \\
\midrule

\multicolumn{12}{l}{\textit{\textbf{Single-Reference Systems}}} \\
\midrule
Llama-ZeroShot & Original & \underline{41.17} & \underline{10.43} & \textbf{19.85} & 46.90 & 11.88 & 22.35 & 40.69 & 10.37 & 19.94 & 83.62 \\
Llama-HighConf & Original & 39.13 & 9.92 & 18.69 & 45.39 & 11.59 & 21.65 & 40.20 & 10.20 & 19.51 & 83.90 \\

\midrule
\multicolumn{12}{l}{\textit{\textbf{Multi-Reference Systems}}} \\
\midrule
Reviewer2 & Original & 29.77 & 7.55 & 15.80 & \textbf{58.62} & \textbf{15.43} & \textbf{32.30} & 22.33 & 5.60 & 11.76 & 83.71 \\
OpenReviewer & Original & 40.92 & \textbf{10.57} & \underline{19.34} & \underline{47.91} & \underline{12.39} & 22.57 & 40.58 & 10.53 & 19.40 & \underline{84.10} \\
Llama-MultiRef & Original & 39.69 & 10.13 & 18.90 & 47.78 & 12.30 & \underline{22.80} & 39.14 & 9.99 & 18.81 & 84.04 \\

\midrule
\multicolumn{12}{l}{\textit{\textbf{Consolidation-Driven Systems}}} \\
\midrule
SEA-E & Original & \textbf{41.32} & 9.26 & 17.07 & 37.56 & 8.37 & 15.30 & \textbf{52.11} & \underline{11.82} & 22.07 & 83.94 \\
AutoRev & Original & 40.21 & 10.08 & 19.03 & 39.09 & 9.74 & 18.22 & 46.42 & 11.78 & \underline{22.45} & 83.62 \\
% Autorev (Re-trained) & Raw & 3.99 & 39.74 & 9.86 & 18.80 & 37.77 & 9.30 & 17.60 & 47.25 & 11.86 & 22.88 & 83.91 \\
Llama-Consolidated & Original & 40.38 & 10.39 & 18.61 & 36.95 & 9.44 & 16.76 & \underline{50.32} & \textbf{13.11} & \textbf{23.76} & \textbf{84.16} \\
% \midrule
% % \multicolumn{12}{l}{\textit{\textbf{Subset Training Experiments}}} \\
% \multicolumn{12}{l}{\textit{\textbf{Ablation Study}}} \\
% \midrule
% Mistral-Consolidated & Original & 41.09 & 8.32 & 16.14 & 36.97 & 7.45 & 14.34 & 52.39 & 10.71 & 21.07 & 83.86 \\
% Mistral-MultiRef & Original & 37.56 & 7.59 & 15.58 & 40.72 & 8.33 & 17.08 & 41.95 & 8.42 & 17.37 & 83.57 \\
% Llama-Consolidated-3k & Original & 39.41 & 7.19 & 15.07 & 35.29 & 6.42 & 13.36 & 50.76 & 9.35 & 19.82 & 83.57 \\
% Llama-MultiRef-3k & Original & 33.84 & 6.10 & 14.49 & 41.61 & 7.65 & 18.07 & 32.88 & 5.85 & 14.01 & 82.87 \\
% Qwen-Consolidated & Original & 37.62 & 6.34 & 14.57 & 35.13 & 5.91 & 13.50 & 45.79 & 7.76 & 18.02 & 83.29 \\
% Qwen-MultiRef & Original & 31.47 & 5.17 & 13.41 & 37.91 & 6.36 & 16.40 & 31.30 & 5.06 & 13.25 & 82.67 \\
\bottomrule
\end{tabular}%
}
\caption{Automatic evaluation metrics comparing model outputs against the average of original unconsolidated target reviews. The highest score in each metric is marked in \textbf{bold}, while the second-highest is \underline{underlined}. BS-F1 denotes BERTScore F1.}
\label{tab:raw_metrics}
\end{table*}
\section{Results}
\label{sec:results}
We compare consolidated and non-consolidated training using ROUGE \citep{lin-2004-rouge}, BERTScore \citep{Zhang*2020BERTScore:} and LongCLIPScore (LCS) \citep{10.1007/978-3-031-72983-6_18} (Section~\ref{main:automatic_eval_results}), supplemented by human evaluation (Section~\ref{main:human_eval_result}).

\begin{table*}[htbp]
\centering
\resizebox{\textwidth}{!}{%
\begin{tabular}{ll ccc ccc ccc c}
\toprule
 & & \multicolumn{3}{c}{\textbf{F1}} & \multicolumn{3}{c}{\textbf{Precision}} & \multicolumn{3}{c}{\textbf{Recall}} & \\
\cmidrule(lr){3-5} \cmidrule(lr){6-8} \cmidrule(lr){9-11}
\textbf{Model} & \textbf{Ground Truth} & \textbf{R-1} & \textbf{R-2} & \textbf{R-L} & \textbf{R-1} & \textbf{R-2} & \textbf{R-L} & \textbf{R-1} & \textbf{R-2} & \textbf{R-L} & \textbf{BS-F1} \\
\midrule

\multicolumn{12}{l}{\textit{\textbf{Single-Reference Systems}}} \\
\midrule
Llama-ZeroShot & Consolidated & 44.98 & 14.85 & 21.26 & \underline{68.62} & 22.74 & \underline{32.54} & 33.87 & 11.16 & 15.99 & 85.48 \\
Llama-HighConf & Consolidated & 41.33 & 12.91 & 19.41 & 63.61 & 20.22 & 30.37 & 32.62 & 10.07 & 15.31 & 85.29 \\

\midrule
\multicolumn{12}{l}{\textit{\textbf{Multi-Reference Systems}}} \\
\midrule
Reviewer2 & Consolidated & 23.96 & 7.57 & 13.23 & \textbf{71.46} & 23.24 & \textbf{41.22} & 14.99 & 4.70 & 8.20 & 84.52 \\
OpenReviewer & Consolidated & 42.63 & 13.56 & 19.85 & 66.84 & 21.44 & 31.55 & 32.44 & 10.27 & 15.02 & 85.58 \\
Llama-MultiRef & Consolidated & 40.32 & 12.53 & 19.04 & 65.33 & 20.64 & 31.48 & 30.70 & 9.45 & 14.42 & 85.34 \\

\midrule
\multicolumn{12}{l}{\textit{\textbf{Consolidation-Driven Systems}}} \\
\midrule
SEA-E & Consolidated & 53.83 & 15.55 & 20.92 & 60.66 & 17.53 & 23.60 & \underline{48.98} & 14.14 & 19.02 & \underline{86.60} \\
AutoRev & Consolidated & \underline{54.26} & \underline{19.78} & \underline{24.42} & 66.90 & \textbf{24.40} & 30.11 & 45.97 & \underline{16.76} & \underline{20.70} & 86.39 \\
% Autorev (Re-trained) & Consolidated & 11.81 & 54.18 & 19.24 & 24.31 & 64.75 & 23.02 & 29.05 & 46.98 & 16.67 & 21.09 & 87.21 \\
Llama-Consolidated & Consolidated & \textbf{57.03} & \textbf{20.90} & \textbf{24.68} & 64.81 & \underline{23.76} & 28.04 & \textbf{51.31} & \textbf{18.80} & \textbf{22.21} & \textbf{87.65} \\
% \midrule
% \multicolumn{12}{l}{\textit{\textbf{Ablation Study}}} \\
% \midrule
% Mistral-Consolidated & Consolidated & 51.36 & 14.72 & 19.58 & 57.49 & 16.51 & 21.94 & 46.87 & 13.42 & 17.86 & 86.62 \\
% Mistral-MultiRef & Consolidated & 38.62 & 9.14 & 15.68 & 54.62 & 13.28 & 22.90 & 32.42 & 7.53 & 12.88 & 84.70 \\
% Llama-Consolidated-3k & Consolidated & 47.70 & 11.94 & 17.83 & 53.20 & 13.34 & 19.91 & 43.79 & 10.94 & 16.34 & 86.10 \\
% Llama-MultiRef-3k & Consolidated & 31.28 & 6.57 & 13.44 & 52.77 & 11.38 & 23.31 & 23.32 & 4.84 & 9.90 & 83.53 \\
% Qwen-Consolidated & Consolidated & 43.51 & 10.12 & 16.74 & 51.57 & 12.02 & 19.88 & 38.07 & 8.84 & 14.62 & 85.57 \\
% Qwen-MultiRef & Consolidated & 29.92 & 5.73 & 12.59 & 49.06 & 9.68 & 21.27 & 22.67 & 4.28 & 9.41 & 83.42 \\
\bottomrule
\end{tabular}%
}
\caption{Automatic evaluation metrics comparing model outputs against the consolidated target reviews. The highest score in each metric is marked in \textbf{bold}, while the second-highest is \underline{underlined}. BS-F1 denotes BERTScore F1.}
\label{tab:consolidated_metrics}
\end{table*}

\begin{table*}[ht]
\centering
\resizebox{\textwidth}{!}{%
\begin{tabular}{ll ccc ccc ccc c}
\toprule
 & & \multicolumn{3}{c}{\textbf{F1}} & \multicolumn{3}{c}{\textbf{Precision}} & \multicolumn{3}{c}{\textbf{Recall}} & \\
\cmidrule(lr){3-5} \cmidrule(lr){6-8} \cmidrule(lr){9-11}
\textbf{Model} & \textbf{Ground Truth} & \textbf{R-1} & \textbf{R-2} & \textbf{R-L} & \textbf{R-1} & \textbf{R-2} & \textbf{R-L} & \textbf{R-1} & \textbf{R-2} & \textbf{R-L} & \textbf{LCS} \\
\midrule
\multicolumn{12}{l}{\textit{\textbf{Single-Reference Systems}}} \\
\midrule
Molmo-SingleRef & Original & 0.403 & 0.156 & 0.366 & 0.450 & 0.176 & 0.408 & 0.401 & 0.156 & 0.365 & 0.812 \\

\midrule
\multicolumn{12}{l}{\textit{\textbf{Multi-Reference Systems}}} \\
\midrule
BLIP-2 OPT-6.7B & Original & 0.398 & 0.150 & 0.356 & 0.413 & 0.155 & 0.367 & 0.414 & 0.157 & 0.372 & \underline{0.816} \\
PaliGemma-3B & Original & \textbf{0.412} & \underline{0.162} & \textbf{0.376} & \textbf{0.469} & \textbf{0.185} & \textbf{0.425} & 0.393 & 0.155 & 0.360 & 0.801 \\
GIT-base & Original & 0.357 & 0.124 & 0.324 & 0.424 & 0.149 & 0.384 & 0.332 & 0.116 & 0.303 & 0.793 \\
Molmo-Multiref & Original & \underline{0.411} & \textbf{0.163} & \underline{0.372} & \underline{0.451} & \underline{0.179} & \underline{0.408} & 0.415 & 0.166 & 0.376 & 0.814 \\

\midrule
\multicolumn{12}{l}{\textit{\textbf{Consolidation-Driven Systems}}} \\
\midrule
Molmo-Consolidated & Original & 0.373 & 0.138 & 0.321 & 0.308 & 0.113 & 0.265 & 0.538 & \underline{0.200} & 0.465 & \textbf{0.839} \\

\midrule
\multicolumn{12}{l}{\textit{\textbf{Long Form Captioners}}} \\
\midrule
LLaVA-1.5-7B & Original & 0.156 & 0.048 & 0.128 & 0.092 & 0.028 & 0.075 & \underline{0.616} & \textbf{0.203} & \underline{0.512} & 0.773 \\
ShareGPT4V-7B & Original & 0.100 & 0.030 & 0.085 & 0.055 & 0.017 & 0.047 & \textbf{0.631} & \textbf{0.203} & \textbf{0.550} & 0.778 \\
Molmo-ZeroShot & Original & 0.134 & 0.035 & 0.106 & 0.076 & 0.020 & 0.060 & 0.664 & 0.184 & 0.538 & 0.765 \\

\bottomrule
\end{tabular}%
}
\caption{Automatic evaluation metrics comparing generated captions against the five original human written captions, averaged over references. LCS denotes LongCLIPScore, which is reference-free. The highest score in each metric is marked in \textbf{bold}, while the second-highest is \underline{underlined}.}
\label{tab:flickr-human-refs}
\end{table*}

\subsection{Automatic Evaluation}
\label{main:automatic_eval_results}

\paragraph{Peer-Review Generation (Tables~\ref{tab:raw_metrics} and \ref{tab:consolidated_metrics}):}
Against the original reviews (Table~\ref{tab:raw_metrics}), consolidation-driven systems reach ROUGE F1 comparable to multi-reference systems, with higher recall and lower precision. This is the expected profile of broader coverage: their longer outputs combine points from several reviewers, so they match any single review less densely while covering more of the panel's points. Since recall against individual reviews cannot by itself distinguish distinct points from generic text, we rely on human evaluation (Section~\ref{main:human_eval_result}) as the direct measure of coverage. Multi-reference systems show the opposite profile: comparatively higher precision and lower recall. On BERTScore, \textit{Llama-Consolidated} is strongest among the main systems. Against the consolidated targets (Table~\ref{tab:consolidated_metrics}), \textit{Llama-Consolidated} is best on both ROUGE F1 and BERTScore, followed by the other consolidation-driven systems. Notably, \textit{Llama-ZeroShot} outperforms \textit{Llama-MultiRef} on both metrics, suggesting that training on conflicting original reviews adds noise rather than teaching synthesis. Example outputs are in Appendix~\ref{app:output-review}.

\begin{table*}[htbp]
\centering
\resizebox{\textwidth}{!}{%
\begin{tabular}{ll ccc ccc ccc c}
\toprule
 & & \multicolumn{3}{c}{\textbf{F1}} & \multicolumn{3}{c}{\textbf{Precision}} & \multicolumn{3}{c}{\textbf{Recall}} & \\
\cmidrule(lr){3-5} \cmidrule(lr){6-8} \cmidrule(lr){9-11}
\textbf{Model} & \textbf{Ground Truth} & \textbf{R-1} & \textbf{R-2} & \textbf{R-L} & \textbf{R-1} & \textbf{R-2} & \textbf{R-L} & \textbf{R-1} & \textbf{R-2} & \textbf{R-L} & \textbf{LCS} \\

\midrule
\multicolumn{12}{l}{\textit{\textbf{Single-Reference Systems}}} \\
\midrule
Molmo-SingleRef & Consolidated & 0.392 & 0.159 & 0.347 & 0.635 & 0.260 & 0.561 & 0.303 & 0.124 & 0.269 & 0.812 \\

\midrule
\multicolumn{12}{l}{\textit{\textbf{Multi-Reference Systems}}} \\
\midrule
GIT-base & Consolidated & 0.334 & 0.115 & 0.293 & 0.591 & 0.207 & 0.517 & 0.245 & 0.084 & 0.216 & 0.793 \\
PaliGemma-3B & Consolidated & 0.393 & 0.152 & 0.345 & \textbf{0.663} & \underline{0.261} & \textbf{0.581} & 0.292 & 0.113 & 0.257 & 0.801 \\
BLIP-2 OPT-6.7B & Consolidated & \underline{0.407} & 0.151 & 0.348 & 0.611 & 0.230 & 0.521 & 0.320 & 0.119 & 0.274 & \underline{0.816} \\
Molmo-Multiref & Consolidated & 0.404 & \underline{0.166} & \underline{0.353} & \underline{0.641} & \textbf{0.265} & \underline{0.561} & 0.315 & 0.131 & 0.276 & 0.814 \\

\midrule
\multicolumn{12}{l}{\textit{\textbf{Consolidation-Driven Systems}}} \\
\midrule
Molmo-Consolidated & Consolidated & \textbf{0.474} & \textbf{0.208} & \textbf{0.391} & 0.512 & 0.225 & 0.422 & 0.478 & \textbf{0.210} & 0.394 & \textbf{0.839} \\

\midrule
\multicolumn{12}{l}{\textit{\textbf{Long Form Captioners}}} \\
\midrule
LLaVA-1.5-7B & Consolidated & 0.248 & 0.078 & 0.181 & 0.164 & 0.051 & 0.119 & 0.571 & 0.190 & 0.422 & 0.773 \\
ShareGPT4V-7B & Consolidated & 0.172 & 0.056 & 0.133 & 0.103 & 0.033 & 0.079 & \underline{0.608} & \underline{0.206} & \textbf{0.479} & 0.778 \\
Molmo-ZeroShot & Consolidated & 0.231 & 0.068 & 0.163 & 0.144 & 0.042 & 0.101 & \textbf{0.649} & 0.198 & \underline{0.468} & 0.765 \\

\bottomrule
\end{tabular}%
}
\caption{Automatic evaluation metrics comparing generated captions against the single consolidated target caption. LCS denotes LongCLIPScore, which is reference-free. The highest score in each metric is marked in \textbf{bold}, while the second-highest is \underline{underlined}.}
\label{tab:flickr-consolidated-refs}
\end{table*}
% \paragraph{Image Captioning (Tables~\ref{tab:flickr-human-refs} and \ref{tab:flickr-consolidated-refs}):}
% On the image captioning part of our dataset, we compare single- and multi-reference captioners (\textit{Molmo-SingleRef}, BLIP-2, GIT, PaliGemma, and our \textit{Molmo-Multiref}), long-form captioners (LLaVA-1.5-7B, ShareGPT4V-7B, and \textit{Molmo-Zeroshot}), and our consolidation-trained \textit{Molmo-Consolidated}. Against the five original captions (Table~\ref{tab:flickr-human-refs}), \textit{Molmo-Consolidated} obtains ROUGE-1 F1 comparable to the single- and multi-reference systems but the highest recall among them at a lower precision. This is the same high-recall, lower-precision pattern that we observed for reviews as a consolidated output covers details spread across several captions, it matches each individual caption less densely while covering more from the reference pool. Notably, the highest recall and lowest precision systems are the long form captioners. Against the consolidated caption (Table~\ref{tab:flickr-consolidated-refs}), \textit{Molmo-Consolidated} leads all systems in ROUGE F1 scores, again with higher recall and lower precision than the single- and multi-reference systems. Over here as well, the long form captioners are amongst the highest recall systems, with the lowest precision.

\paragraph{Image Captioning (Tables~\ref{tab:flickr-human-refs} and \ref{tab:flickr-consolidated-refs}):}
We compare single- and multi-reference captioners (\textit{Molmo-SingleRef}, BLIP-2, GIT, PaliGemma, \textit{Molmo-Multiref}), long-form captioners (LLaVA-1.5-7B, ShareGPT4V-7B, \textit{Molmo-Zeroshot}), and \textit{Molmo-Consolidated}. Against the consolidated caption (Table~\ref{tab:flickr-consolidated-refs}), \textit{Molmo-Consolidated} leads all systems in ROUGE F1. Against the five original captions (Table~\ref{tab:flickr-human-refs}), its F1 is below that of the best single- and multi-reference systems. In both cases it shows higher recall and lower precision than these systems, the same pattern as for reviews: its longer outputs (Appendix~\ref{sec:appendix_length_analysis}) cover details spread across several captions. The long-form captioners are the extreme case, with the highest recall and lowest precision. Example outputs are in Appendix~\ref{app:output-caption}.

% As in peer-review generation, we first report ROUGE against both the original and the consolidated references.

% We complement our ROUGE metrics with LongCLIPScore (LCS) \citep{10.1007/978-3-031-72983-6_18}, which scores image-text alignment directly from the image and needs no reference. LCS is therefore independent of the ground truth used and gives identical scores in both tables, which makes it the most trustworthy yardstick for comparing the systems. Under LCS, \textit{Molmo-Consolidated} attains the best score ($0.839$), ahead of \textit{Molmo-SingleRef} ($0.812$) and all multi-reference captioners ($0.793$--$0.816$). This confirms that the gain from consolidation is not an artifact of how the references are chosen. The long-form captioners do not work well on this task: \textit{LLaVA-1.5-7B}, \textit{ShareGPT4V-7B} and \textit{Molmo-Zeroshot} obtain the lowest LCS of all systems ($0.773$, $0.778$ and $0.765$ respectively).

We complement ROUGE with LongCLIPScore (LCS), which scores image--text alignment directly from the image without references, so it is identical across both tables and independent of the target used. \textit{Molmo-Consolidated} attains the best LCS, ahead of \textit{Molmo-SingleRef} and all multi-reference captioners. The long-form captioners obtain the lowest LCS of all systems. Because LCS uses no reference captions, this advantage does not depend on whether outputs are scored against the original or the consolidated references.

\subsection{Ablation Across Model Families}
\label{sec:ablation}
% Many of the existing baselines evaluated for the task of Peer-Review Generation utilize \textit{Llama-3.1-8B-Instruct} as their backbone model. To generalize our claims across different model families, we conduct an ablation study applying the methodology detailed in Section~\ref{sec:methodology} to two other backbones from different model families. 
While many of the existing baselines evaluated for the task of Peer-Review Generation use \textit{Llama-3.1-8B-Instruct} as their base model, we go a step further to test whether the benefits extend beyond one architecture. To test that our claims generalize, we extend our methodology (Section~\ref{sec:methodology}) to two additional, distinct base models: \textit{Mistral-7B-Instruct-v0.3} and \textit{Qwen2.5-7B-Instruct}. 

As reported in  Tables~\ref{tab:ablation_original_metrics} and \ref{tab:ablation_consolidated_metrics}, \textit{Mistral-Consolidated} and \textit{Qwen-Consolidated} parallel the improvements observed in our Llama-based variant. Against the original references, the consolidated models achieve higher ROUGE F1, ROUGE Recall, and BERTScore F1, but lower ROUGE Precision than their MultiRef counterparts, a trend also observed in Section~\ref{main:automatic_eval_results}. Against the consolidated references, the consolidated models achieve higher scores across all metrics except ROUGE-L Precision, where the MultiRef variants perform better.

% While \textit{Mistral-Consolidated} achieves a higher BERTScore than the other model families, \textit{Llama-Consolidated} remains highly competitive and performs comparably across all evaluated metrics. When evaluating against the original reviews, consolidation-driven systems consistently yield higher ROUGE-Recall than the multi-reference trained models while maintaining comparable ROUGE-F1 scores. This signifies that their outputs successfully cover a broader range of valid critiques spanning multiple reviews for a single paper. Furthermore, when evaluating against the consolidated target reviews, consolidation-driven models across all three families achieve strictly higher scores in both the ROUGE and BERTScore metrics compared to the multi-reference baselines. This demonstrates that exposing the model directly to multiple, diverse original peer reviews during training induces noise that hinders effective learning.
\subsection{Statistical Significance Tests}
\label{sec:statistical_tests}
To test whether differences in performance are statistically significant, we conduct paired statistical significance testing across all evaluations. We employ the paired Student's t-test \citep{student1908probable} metric for BERTScore and LCS and paired bootstrap resampling (10,000 iterations) for ROUGE metrics.
% where ** denotes highly statistically significant differences ($p < 0.01$). 

Across both Peer-Review Generation and Image Captioning, Llama-Consolidated and Molmo-Consolidated demonstrate statistically significant differences compared to the baselines against both original and consolidated targets ($p < 0.01$). Furthermore, the effects of consolidation-driven training generalize across three model families, maintaining significant statistical performance gains ($p < 0.01$). Detailed statistical results are provided in Appendix~\ref{app:stats_results}.

\subsection{Human Evaluation}
\label{main:human_eval_result}
\begin{table}[htbp]
\centering
\begin{tabular}{lcccc}
\toprule
\textbf{Section} & \textbf{LC} & \textbf{LZ} & \textbf{LH} & \textbf{LM} \\
\midrule
Summary    & \textbf{2.08} & \underline{2.50} & 2.57 & 2.85 \\
Strengths  & \textbf{1.53} & \underline{2.33} & 2.98 & 3.16 \\
Weaknesses & \textbf{1.71} & 2.81 & 2.78 & \underline{2.70} \\
Questions  & \textbf{1.48} & \underline{2.40} & 2.92 & 3.20 \\
\midrule
\textbf{Overall} & \textbf{1.70} & \underline{2.51} & 2.81 & 2.98 \\
\bottomrule
\end{tabular}
    \caption{Average human-assigned rank per section (1-4, lower is better) for reviews generated by \textit{Llama-Consolidated} (LC), \textit{Llama-ZeroShot} (LZ), \textit{Llama-HighConf} (LH), \textit{Llama-MultiRef} (LM). The best-performing result is highlighted in \textbf{bold}, and the second-best result is \underline{underlined}.}
\label{tab:human_rankings_review}
\end{table}
% For peer-review generation, \textit{Llama-Consolidated} achieves a superior average rank of $\mathbf{1.70}$ across all sections (Table~\ref{tab:human_rankings_review}). As shown in Figure~\ref{fig:avg_rank}, this advantage is most pronounced in the \textit{Questions} and \textit{Strengths} sections ($1.53$ vs.\ $3.16$), demonstrating its ability to capture diverse reviewing perspectives. Supported by pairwise comparisons (Figure~\ref{fig:win_matrix}), these results empirically confirm that consolidated-reference training produces measurably more comprehensive reviews than single- or multi-reference approaches. We obtain a \textbf{Kendall's} $ \mathbf{W = 0.614}$ overall, which constitutes substantial agreement \citep{Landis1977TheMO}.

For peer-review generation, \textit{Llama-Consolidated} achieves the best average rank of $\mathbf{1.70}$ across all sections (Table~\ref{tab:human_rankings_review}). As shown in Figure~\ref{fig:avg_rank}, the advantage of \textit{Llama-Consolidated} against \textit{Llama-MultiRef} is largest in \textit{Questions} ($1.48$ vs.\ $3.20$) and \textit{Strengths} ($1.53$ vs.\ $3.16$), suggesting that it covers a broader range of reviewer perspectives. Pairwise comparisons (Figure~\ref{fig:win_matrix}) point the same way.
\begin{table}[htbp]
\centering
\begin{tabular}{lc}
\toprule
\textbf{Model} & \textbf{Average Rank} \\
\midrule
Molmo-SingleRef   & 2.68  \\
Molmo-MultiRef    & \underline{2.64} \\
Molmo-Consolidated & \textbf{1.87}  \\
Molmo-ZeroShot    & 2.82  \\
\bottomrule
\end{tabular}
\caption{Average human-assigned rank (1-4, lower is better) for captions generated by \textit{Molmo-Consolidated}, \textit{Molmo-ZeroShot}, \textit{Molmo-SingleRef}, \textit{Molmo-MultiRef}. The best-performing result is highlighted in \textbf{bold} and the second-best result is \underline{underlined}.}
\label{tab:human_ranking_captions}
\end{table}
% For image captioning, \textit{Molmo-Consolidated} achieves the best average human rank (Table~\ref{tab:human_ranking_captions}) and decisively outperforms all baselines. In pairwise comparisons (Figure~\ref{fig:win_matrix_image}) while the alternative paradigms fail to establish clear dominance over one another, the significant gap achieved by \textit{Molmo-Consolidated} confirms that reference consolidation yields measurably more accurate and relevant captions.

For image captioning, \textit{Molmo-Consolidated} achieves the best average rank of 1.87, outperforming \textit{Molmo-MultiRef} (2.64), \textit{Molmo-SingleRef} (2.68), and \textit{Molmo-ZeroShot} (2.82) (Table~\ref{tab:human_ranking_captions}). In pairwise comparisons (Figure~\ref{fig:win_matrix_image}), the non-consolidated variants show no clear ordering between one another, while \textit{Molmo-Consolidated} is ranked above each of them, suggesting that annotators overall judge consolidated captions to be more accurate and natural. The detailed results are reported and discussed in Appendix~\ref{app:human_eval}. 

\section{Conclusion}
\label{sec:conclusion}

We investigate training target selection for peer-review generation and image captioning, comparing single-reference, multi-reference, and consolidation-driven training. To support this, we introduce \textbf{MERC-67K}. Across both tasks, consolidation-driven models produce better outputs (higher recall for reviews, better LongCLIPScore for captions). Human annotators rank them first, for coverage in reviews and overall quality in captions. Unconsolidated multi-reference training also often underperforms zero-shot inference (for peer-review generation), suggesting that unaligned references may add noise. Together, these findings provide empirical support for target consolidation in the intermediary NLG tasks we study.

\section*{Limitations}
\label{sec:limitations}

\paragraph{Scope:} Our study covers two intermediary-entropy tasks. The peer-review data come from machine learning venues (ICLR, NeurIPS, COLM), so extending the findings to other disciplines with different reviewing norms is left to future work. The captioning experiments use \textit{Molmo-7B-O}, the only suitable base model whose training explicitly excludes any open source dataset. This precludes a cross-model ablation for that task.

\paragraph{Models and training:} Experiments use models of up to 8B parameters with supervised LoRA fine-tuning. The cross-architecture ablation on \textit{Mistral-7B} and \textit{Qwen2.5-7B} uses a 3,000-paper subset. Whether larger models or other finetuning regimes behave similarly remains open.

\paragraph{Consolidation Method:} We use one consolidation procedure per task (SEA-S for reviews and a prompted \textit{Llama-3.1-8B-Instruct} for captions). Our results therefore show the benefit of consolidation under these procedures. Consolidation does not verify the correctness of individual review points, so errors that do not conflict with other reviews may carry into the target. Comparing alternative strategies, designing a new consolidation method, or weighting reviews by quality is left for future work.

\paragraph{Evaluation:} Our human evaluation of reviews measures coverage of the original reviewing panel, and the caption study is modest in size. Assessing factual grounding and usefulness to authors, and validating consolidated targets more extensively, would further strengthen these results.

\bibliography{custom}

\appendix
\section{Dataset Statistics for Entropy Spectrum Analysis}
\label{app:entropy-dataset}
To map the NLG entropy spectrum, we evaluate six generation tasks using established multi-reference test benchmarks: Machine Translation (English--German) from \citet{ea-mt-benchmark, conia-etal-2024-towards}, Paraphrase Generation from \citet{10.1007/978-3-030-77385-4_36}, Peer-Review Generation from our MERC-67K test split, Image Captioning from \citet{7410660}, Story Generation from \citet{fan-etal-2018-hierarchical}, and Dialogue Generation from \citet{sai-etal-2020-improving}. Table~\ref{tab:dataset_entropy_stats} summarizes the number of test instances and the median reference count per sample across each task.

\begin{table}[htbp]
\centering
\small
\begin{tabular}{lcc}
\toprule
\textbf{Task} & \textbf{Test Samples} & \textbf{\shortstack{Median Refs /\\Sample}} \\
\midrule
Translation            & 3,003 & 2 \\
Paraphrase Generation  & 3,246 & 3 \\
Peer-Review Generation & 3,659 & 4\\
Image Captioning       & 1,000 & 5 \\
Story Generation       & 2,761 & 3 \\
Dialogue Generation    & 1,142 & 5 \\
\bottomrule
\end{tabular}
\caption{Dataset statistics and median reference counts across the six evaluated NLG tasks used to compute empirical von Neumann entropy.}
\label{tab:dataset_entropy_stats}
\end{table}

\section{MERC67K Dataset: Supplementary Details}
\subsection{Prompt for Image Caption Consolidation}
\label{app:caption-prompt}
To synthesize the five human-written references per image in Flickr30k into a single consolidated target for MERC-67K, we prompt \textit{Llama-3.1-8B-Instruct} using Prompt~\ref{lst:caption-consolidate-prompt}. Prompting is conducted purely over text via the standard chat template without visual input, producing the target captions used to train \textit{Molmo-Consolidated} and evaluate other baselines for this task.
\begin{tcolorbox}[title={Prompt Used for Caption Consolidation}, width=\linewidth, colback=white, colframe=gray, arc=0pt, outer arc=5pt, boxrule=0.5pt, leftrule=2pt, rightrule=2pt, right=4pt, left=4pt, top=4pt, bottom=4pt, toprule=0pt, bottomrule=2pt]
\label{lst:caption-consolidate-prompt}
\textbf{System:} \\
You consolidate multiple human image captions into a single high-quality caption. Preserve all non-conflicting details from the sources. Do not invent objects, attributes, or actions that none of the captions mention. If captions disagree, prefer the detail supported by more captions or the more specific phrasing. Output ONLY the consolidated caption, no quotes, no preamble, no bullet points.\\
\textbf{User:} \\
Human captions for the same image:
\begin{enumerate}
\item \textit{<caption$_1$>}
\item \textit{<caption$_2$>}
\item \textit{<caption$_3$>}
\item \textit{<caption$_4$>}
\item \textit{<caption$_5$>}
\end{enumerate}
Write one consolidated caption:
\end{tcolorbox}
\section{Methodology: Supplementary Details}
\subsection{Training Setup}
\label{sec:appendix-imp}
We fine-tune our base models for both tasks using four NVIDIA RTX A6000 GPUs (48GB VRAM each).

We use the prompts mentioned below across all the models trained for peer-review generation and image captioning respectively. We apply Low-Rank Adaptation (LoRA) with a rank of 8 and a scaling factor ($\alpha = 16$). For the peer-review generation task, LoRA is integrated into the following transformer layers of Llama: \textit{q\_proj}, \textit{k\_proj}, \textit{v\_proj}, \textit{o\_proj}, \textit{gate\_proj}, \textit{up\_proj}, and \textit{down\_proj}. For the image captioning task, LoRA is applied to Molmo's language, vision, and projector modules: \textit{att\_proj}, \textit{ff\_proj}, \textit{attn\_out}, and \textit{ff\_out} in the OLMo language model; \textit{attention.wq}, \textit{attention.wk}, \textit{attention.wv}, \textit{attention.wo}, \textit{feed\_forward.w1}, and \textit{feed\_forward.w2} in the CLIP vision backbone; and \textit{w1}, \textit{w2}, and \textit{w3} in the image projector. Models are fine-tuned for 3 epochs. The full set of hyperparameters used during training is listed in Table~\ref{tab:hyperparams}.
\begin{tcolorbox}[title={Prompt Used During Finetuning (Peer-Review Generation)}, width=\linewidth, colback=white, colframe=gray, arc=0pt, outer arc=5pt, boxrule=0.5pt, leftrule=2pt, rightrule=2pt, right=4pt, left=4pt, top=4pt, bottom=4pt, toprule=0pt, bottomrule=2pt]
\label{lst:prompt-review}
Generate a structured review for the research paper passages provided below. The review should include a summary of the paper, its strengths, weaknesses, and questions for the authors.

\vspace{0.5em}
\textbf{\#\#\# Research Paper Passages:} \\
\textbf{\#\#\# Review for the paper:} \\
\textbf{**Summary**} \\
\textbf{**Strengths**} \\
\textbf{**Weaknesses**} \\
\textbf{**Questions**} \\
\textit{<EOS\_TOKEN>}
\end{tcolorbox}

\begin{table}[htbp]
% \small
\centering
\begin{tabular}{ll}
\toprule
\textbf{Hyperparameter} & \textbf{Value} \\
\midrule
% Maximum Sequence Length & 6000 \\
Batch Size & 1 \\
Gradient Accumulation Steps & 4 \\
16-bit Floating Point Precision & True \\
Optimizer & adamw\_torch \\
Learning Rate & $1 \times 10^{-4}$ \\
Max Gradient Norm & 0.3 \\
Warmup Ratio & 0.03 \\
LoRA Rank & 8 \\
LoRA Scaling Factor & 16 \\
LoRA Dropout & 0.05 \\
Weight Decay & 0.01 \\
LR Scheduler Type & cosine \\
\bottomrule
\end{tabular}
\caption{Hyperparameters Used During Finetuning}
\label{tab:hyperparams}
\end{table}
For all existing systems, we strictly adhere to the hyperparameter, prompt, and inference settings reported in their original publications during evaluation.
% GPU usage
% Prompt
% Hyperparams
\begin{tcolorbox}[title={Prompt Used During Finetuning (Image Caption Generation)}, width=\linewidth, colback=white, colframe=gray, arc=0pt, outer arc=5pt, boxrule=0.5pt, leftrule=2pt, rightrule=2pt, right=4pt, left=4pt, top=4pt, bottom=4pt, toprule=0pt, bottomrule=2pt]
\label{lst:prompt-caption}
Describe this image.
\vspace{0.5em}
\textbf{\#\#\# Image:} \\
\textit{<IMAGE>} \\
\textbf{\#\#\# Caption:} \\
\textit{<CAPTION>} \\
\textit{<EOS\_TOKEN>}
\end{tcolorbox}
% The prompt shown above is used to fine-tune the base \textit{Llama-3.1-8B-Instruct} model to develop the the \textit{Llama-HighConf}, \textit{Llama-MultiRef}, \textit{Llama-Consolidated} variants, as well as for zero-shot inference with \textit{Llama-Inference}. All other existing systems (\textit{Reviewer2}, \textit{OpenReviewer}, \textit{SEA-E}, and \textit{AutoRev}) are evaluated using the specific system prompts detailed in their respective papers.

\subsection{Inference Details}
\label{sec:inference-details}
After finetuning, LoRA adapters are merged into the corresponding base weights.
We then generate outputs on the held-out test split using the same task prompts as during training
(Prompt~\ref{lst:prompt-review} and~\ref{lst:prompt-caption}).
Unless noted otherwise, decoding uses nucleus sampling with temperature $0.7$,
$\mathrm{top}\text{-}p=0.9$, and repetition penalty $1.15$.
\paragraph{Peer-Review Generation:}
Reviews are generated with \textsc{vLLM} on the merged base model + LoRA checkpoints in \texttt{bfloat16}.
Each test paper is converted to markdown and truncated to the main body of the paper;
prompts are rendered with the model's chat template, with left-side truncation applied when the context budget is exceeded
(maximum sequence length $16{,}384$ tokens).
We sample up to $2{,}048$ new tokens with an additional frequency penalty of $0.1$,
stopping at the native end-of-turn / EOS tokens.
The same settings are used across all the model variants.

\paragraph{Image Captioning:}
Captions are generated with Hugging Face Transformers on merged Molmo-7B-O checkpoints in
\texttt{float16}, using Molmo's native multi-crop visual encoding ($12$ crops) and
\texttt{generate\_from\_batch} with KV caching enabled.
The instruction is \textit{``Describe this image.''}
We stop at EOS / \texttt{<|endoftext|>}.
The same decoding settings are used across all model variants.

For all other standard models, the inference settings were directly borrowed from their literature. 
% ; only the underlying weights differ.

% \label{sec:appendix_data_split}
% \begin{table*}[htbp]
% \centering
% \begin{tabular}{lcccc}
% \toprule
% \textbf{Conference} & \textbf{Train} & \textbf{Validation} & \textbf{Test} & \textbf{Total} \\ \midrule
% COLM 2024           & 234            & 29                  & 30            & 293            \\
% COLM 2025           & 324            & 40                  & 41            & 405            \\
% NeurIPS 2023        & 2686           & 336                 & 336           & 3358           \\
% NeurIPS 2024        & 3319           & 415                 & 415           & 4149           \\
% ICLR 2024           & 4507           & 564                 & 563           & 5634           \\
% ICLR 2025           & 6610           & 827                 & 831           & 8268           \\
% ICLR 2026           & 11544          & 1443                & 1443          & 14430          \\ \midrule
% \textbf{Total}      & \textbf{29224} & \textbf{3654}       & \textbf{3659} & \textbf{36537} \\ \bottomrule
% \end{tabular}
% \caption{Train, Validation, and Test distribution of papers across all conferences in MERC-67K.}
% \label{tab:split-stats}
% \end{table*}
% We partition the MERC-67K dataset into training (80\%), validation (10\%), and testing (10\%) subsets. Table~\ref{tab:split-stats} provides the statistics of the number of papers allocated to each split across various target conferences, ensuring a consistent proportional distribution across all venues.

\section{Output Length Analysis}
\label{sec:appendix_length_analysis}
To better understand the impact of target formulation on model verbosity, we analyze the token length distributions of both the ground truth targets and the generated outputs across all evaluated paradigms. 

For the peer-review generation task, the dataset contains an average of 3.98 reviews per paper with a median of 4. As detailed in Table \ref{tab:review_lengths}, the consolidation process naturally yields longer targets, with the Consolidated References (CR) averaging 1327.18 tokens compared to 578.07 tokens for raw individual reviews. Consequently, the fine-tuned \textit{Llama-Consolidated} model generates longer outputs than both \textit{Llama-HighConf} and \textit{Llama-MultiRef}. 

\begin{table}[ht]
\centering
\small
\begin{tabular}{lrrr}
\toprule
\textbf{Set} & \textbf{n} & \textbf{Mean Tok} & \textbf{Median Tok} \\
\midrule
Raw Reviews & 145,530 & 578.07 & 505\\
CR & 36,537 & 1327.18 & 1272\\
Consolidated & 3,659 & 704.67 & 702\\
Highconf & 3,659 & 593.57 & 465\\
Multiref & 3,659 & 508.48 & 430\\
ZeroShot & 3,659 & 435.47 & 421\\
\bottomrule
\end{tabular}
\caption{Output length statistics for peer-review generation (measured in \textit{Llama-3.1-8B-Instruct} tokens).}
\label{tab:review_lengths}
\end{table}

A similar trend is observed in the image captioning task, which features an average of 5.00 captions per image with a median of 5. Table \ref{tab:caption_lengths} shows that the Consolidated References (CR) averages 27.88 tokens, roughly double the length of the individual human references (mean 13.97 tokens). \textit{Molmo-Consolidated} closely mirrors this distribution with an average generation length of 25.20 tokens, while \textit{Molmo-MultiRef} and \textit{Molmo-SingleRef} generate shorter outputs averaging 12.36 and 12.07 tokens, respectively. Notably, the zero-shot baseline generates highly verbose captions averaging 127.68 tokens, indicating a failure to align with the task's expected density without fine-tuning.

\begin{table}[ht]
\centering
\small
\begin{tabular}{lrrr}
\toprule
\textbf{Set} & \textbf{n} & \textbf{Mean Tok} & \textbf{Median Tok} \\
\midrule
Raw Captions & 155,070 & 13.97 & 13\\
CR & 31,014 & 27.88 & 27\\
Consolidated & 1,000 & 25.20 & 24\\
Singleref & 1,000 & 12.07 & 11\\
Multiref & 1,000 & 12.36 & 12\\
Zeroshot & 1,000 & 127.68 & 128\\
\bottomrule
\end{tabular}
\caption{Output length statistics for image captioning (measured in \textit{Molmo-7B-O} tokens).}
\label{tab:caption_lengths}
\end{table}

\section{Ablation Across Model Families}
\label{app:ablation}
We provide the quantitative results for the cross-architecture ablation study discussed in Section~\ref{sec:ablation}. Following the methodology outlined in Section~\ref{sec:methodology}, we extended our training paradigms to two additional, distinct backbone models: \textit{Mistral-7B-Instruct-v0.3} and \textit{Qwen2.5-7B-Instruct}. Detailed performance metrics evaluated against the original and consolidated ground-truth reviews are reported in Tables~\ref{tab:ablation_original_metrics} and \ref{tab:ablation_consolidated_metrics}, respectively.
\begin{table*}[htbp]
\centering
\resizebox{\textwidth}{!}{%
\begin{tabular}{ll ccc ccc ccc c}
\toprule
 & & \multicolumn{3}{c}{\textbf{F1}} & \multicolumn{3}{c}{\textbf{Precision}} & \multicolumn{3}{c}{\textbf{Recall}} & \\
\cmidrule(lr){3-5} \cmidrule(lr){6-8} \cmidrule(lr){9-11}
\textbf{Model} & \textbf{Ground Truth} & \textbf{R-1} & \textbf{R-2} & \textbf{R-L} & \textbf{R-1} & \textbf{R-2} & \textbf{R-L} & \textbf{R-1} & \textbf{R-2} & \textbf{R-L} & \textbf{BS-F1} \\
\midrule
\multicolumn{12}{l}{\textit{\textbf{Mistral-7B}}} \\
\midrule
Mistral-Consolidated & Original & \textbf{41.09} & \textbf{8.32} & \textbf{16.14} & 36.97 & 7.45 & 14.34 & \textbf{52.39} & \textbf{10.71} & \textbf{21.07} & \textbf{83.86} \\
Mistral-MultiRef & Original & 37.56 & 7.59 & 15.58 & \textbf{40.72} & \textbf{8.33} & \textbf{17.08} & 41.95 & 8.42 & 17.37 & 83.57 \\

\midrule
\multicolumn{12}{l}{\textit{\textbf{Llama-3.1-8B}}} \\
\midrule
Llama-Consolidated-3k & Original & \textbf{39.41} & \textbf{7.19} & \textbf{15.07} & 35.29 & 6.42 & 13.36 & \textbf{50.76} & \textbf{9.35} & \textbf{19.82} & \textbf{83.57} \\
Llama-MultiRef-3k & Original & 33.84 & 6.10 & 14.49 & \textbf{41.61} & \textbf{7.65} & \textbf{18.07} & 32.88 & 5.85 & 14.01 & 82.87 \\

\midrule
\multicolumn{12}{l}{\textit{\textbf{Qwen-2.5-7B}}} \\
\midrule
Qwen-Consolidated & Original & \textbf{37.62} & \textbf{6.34} & \textbf{14.57} & 35.13 & 5.91 & 13.50 & \textbf{45.79} & \textbf{7.76} & \textbf{18.02} & \textbf{83.29} \\
Qwen-MultiRef & Original & 31.47 & 5.17 & 13.41 & \textbf{37.91} & \textbf{6.36} & \textbf{16.40} & 31.30 & 5.06 & 13.25 & 82.67 \\
\bottomrule
\end{tabular}%
}
\caption{Ablation study across different model architectures evaluated against original unconsolidated target reviews. Bold values indicate the superior performance between the consolidation-driven and multi-reference paradigm for each respective model family. BS-F1 denotes BERTScore F1.}
\label{tab:ablation_original_metrics}
\end{table*}

\begin{table*}[htbp]
\centering
\resizebox{\textwidth}{!}{%
\begin{tabular}{ll ccc ccc ccc c}
\toprule
 & & \multicolumn{3}{c}{\textbf{F1}} & \multicolumn{3}{c}{\textbf{Precision}} & \multicolumn{3}{c}{\textbf{Recall}} & \\
\cmidrule(lr){3-5} \cmidrule(lr){6-8} \cmidrule(lr){9-11}
\textbf{Model} & \textbf{Ground Truth} & \textbf{R-1} & \textbf{R-2} & \textbf{R-L} & \textbf{R-1} & \textbf{R-2} & \textbf{R-L} & \textbf{R-1} & \textbf{R-2} & \textbf{R-L} & \textbf{BS-F1} \\
\midrule
\multicolumn{12}{l}{\textit{\textbf{Mistral-7B}}} \\
\midrule
Mistral-Consolidated & Consolidated & \textbf{51.36} & \textbf{14.72} & \textbf{19.58} & \textbf{57.49} & \textbf{16.51} & 21.94 & \textbf{46.87} & \textbf{13.42} & \textbf{17.86} & \textbf{86.62} \\
Mistral-MultiRef & Consolidated & 38.62 & 9.14 & 15.68 & 54.62 & 13.28 & \textbf{22.90} & 32.42 & 7.53 & 12.88 & 84.70 \\

\midrule
\multicolumn{12}{l}{\textit{\textbf{Llama-3.1-8B }}} \\
\midrule
Llama-Consolidated-3k & Consolidated & \textbf{47.70} & \textbf{11.94} & \textbf{17.83} & \textbf{53.20} & \textbf{13.34} & 19.91 & \textbf{43.79} & \textbf{10.94} & \textbf{16.34} & \textbf{86.10} \\
Llama-MultiRef-3k & Consolidated & 31.28 & 6.57 & 13.44 & 52.77 & 11.38 & \textbf{23.31} & 23.32 & 4.84 & 9.90 & 83.53 \\

\midrule
\multicolumn{12}{l}{\textit{\textbf{Qwen-2.5-7B}}} \\
\midrule
Qwen-Consolidated & Consolidated & \textbf{43.51} & \textbf{10.12} & \textbf{16.74} & \textbf{51.57} & \textbf{12.02} & 19.88 & \textbf{38.07} & \textbf{8.84} & \textbf{14.62} & \textbf{85.57} \\
Qwen-MultiRef & Consolidated & 29.92 & 5.73 & 12.59 & 49.06 & 9.68 & \textbf{21.27} & 22.67 & 4.28 & 9.41 & 83.42 \\
\bottomrule
\end{tabular}%
}
\caption{Ablation study across different model architectures evaluated against the consolidated target reviews. Bold values indicate the superior performance between the consolidation-driven and multi-reference paradigm for each respective model family. BS-F1 denotes BERTScore F1.}
\label{tab:ablation_consolidated_metrics}
\end{table*}
\section{Model Outputs}
\subsection{Peer-Review Generation}
\label{app:output-review}
% ==========================================
% Card 1: Orange
% ==========================================
\begin{tcolorbox}[reviewcard, colback=boxorange]
    \modelbadge{titleorange}{Llama-ZeroShot}
    
    \textbf{Summary:} \\
    The paper proposes a novel method, Unified Framework for Heterogeneous Semi-supervised Learning (Uni-HSSL), to address the Heterogeneous Semi-Supervised Learning (HSSL) problem. It learns a fine-grained classification model cohesively under a unified framework by amalgamating labeled and unlabeled class categories. The framework relies on pseudo-labeling, cross-domain prototype alignment, and progressive inter-domain mixup.
    
    \vspace{1.5mm}
    \textbf{Strengths:}
    \begin{itemize}[leftmargin=*, nosep]
        \item Proposes a novel method to address the challenging Heterogeneous Semi-Supervised Learning (HSSL) problem.
        \item Consists of three well-designed technical components to tackle HSSL challenges.
        \item Outperforms state-of-the-art semi-supervised learning and unsupervised domain adaptation methods on benchmark datasets.
    \end{itemize}
    
    \vspace{1.5mm}
    \textbf{Weaknesses:}
    \begin{itemize}[leftmargin=*, nosep]
        \item Assumes labeled and unlabeled domains share the same semantic classes, which may not always hold in real-world scenarios.
        \item Requires a large amount of labeled data to pre-train the feature encoder and the classifier.
        \item May struggle when labeled and unlabeled domains have very different distributions.
    \end{itemize}
    
    \vspace{1.5mm}
    \textbf{Questions:} \\
    How does the method handle cases when the labeled and unlabeled domains do not share the same set of semantic classes, and can it be extended to multi-task or transfer learning scenarios?
\end{tcolorbox}

% ==========================================
% Card 2: Purple
% ==========================================
\begin{tcolorbox}[reviewcard, colback=boxpurple]
    \modelbadge{titlepurple}{Llama-Highconf}
    
    \textbf{Summary:} \\
    This paper presents heterogeneous semi-supervised learning, a combination of semi-supervised learning and unsupervised domain adaptation. To address differing label and class feature distributions across domains, the authors proposed a unified framework comprising pseudo-labeling, cross-domain prototype alignment, and progressive inter-domain mixup.
    
    \vspace{1.5mm}
    \textbf{Strengths:}
    \begin{itemize}[leftmargin=*, nosep]
        \item Proposes a new task, HSSL, which is a combination of semi-supervised learning and unsupervised domain adaptation.
        \item The paper is well-written and easy to follow.
        \item The proposed method achieves better performance than the state-of-the-art methods.
    \end{itemize}
    
    \vspace{1.5mm}
    \textbf{Weaknesses:}
    \begin{itemize}[leftmargin=*, nosep]
        \item The assumption that labeled and unlabeled data have different label and class feature distributions is not discussed in detail, and there is no evaluation on the effect of the domain shift.
        \item Experimental results are not fully convincing; in some cases, the proposed method's performance is lower than supervised baselines.
    \end{itemize}
    
    \vspace{1.5mm}
    \textbf{Questions:} \\
    Please see the weaknesses.
\end{tcolorbox}

% ==========================================
% Card 4: Blue (Flows automatically to next column/page)
% ==========================================
\begin{tcolorbox}[reviewcard, colback=boxblue]
    \modelbadge{titleblue}{Llama-Multiref}
    
    \textbf{Summary:} \\
    This paper introduces the Heterogeneous Semi-Supervised Learning (HSSL) setup, which bridges SSL and unsupervised domain adaptation. It aims to learn a prediction model using labeled and unlabeled data from heterogeneous domains. The authors propose the Uni-HSSL framework to adaptively handle inter-domain heterogeneity and leverage semantic relationships.
    
    \vspace{1.5mm}
    \textbf{Strengths:}
    \begin{itemize}[leftmargin=*, nosep]
        \item Introduces a novel problem setup termed as HSSL, presenting unique challenges by bridging SSL and UDA.
    \end{itemize}
    
    \vspace{1.5mm}
    \textbf{Weaknesses:}
    \begin{itemize}[leftmargin=*, nosep]
        \item The paper is poorly written, making it difficult to understand the core idea.
        \item Does not provide a theoretical proof of the proposed method.
        \item The innovation is weak, consisting of a combination of several existing methods, and experimental results/baselines are insufficient.
    \end{itemize}
    
    \vspace{1.5mm}
    \textbf{Questions:} \\
    See the weakness.
\end{tcolorbox}
% ==========================================
% Card 3: Green
% ==========================================
\begin{tcolorbox}[reviewcard, colback=boxgreen]
    \modelbadge{titlegreen}{Llama-Consolidated}
    
    \textbf{Summary:} \\
    The paper introduces a novel framework for heterogeneous semi-supervised learning (HSSL). Uni-HSSL integrates pseudo-labeling, prototype alignment, and inter-domain mixup to align class prototypes and generate synthetic training data. While effective, the paper is criticized for its lack of novelty and insufficient comparison with recent state-of-the-art methods.
    
    \vspace{1.5mm}
    \textbf{Strengths:}
    \begin{itemize}[leftmargin=*, nosep]
        \item The problem setup is well-defined, addressing a novel and important real-world problem.
        \item Integrates innovative and effective components to form the Uni-HSSL method.
        \item Experimental results demonstrate improvements over existing methods, supported by a detailed ablation study.
    \end{itemize}
    
    \vspace{1.5mm}
    \textbf{Weaknesses:}
    \begin{itemize}[leftmargin=*, nosep]
        \item Lacks novelty in its components, as many techniques are already widely used in SSL and domain adaptation.
        \item Does not adequately compare the proposed method with recent state-of-the-art methods.
        \item Suffers from typographical errors and lacks clarity in certain sections.
    \end{itemize}
    
    \vspace{1.5mm}
    \textbf{Questions:} \\
    Can the authors clarify the motivation behind the problem setup, explain why the proposed method is more effective than existing methods, and provide a more detailed comparison with recent state-of-the-art methods?
\end{tcolorbox}
We present reviews generated for a randomly selected test paper, \textit{A Unified Framework for Heterogeneous Semi-supervised Learning} (OpenReview: \href{https://openreview.net/forum?id=uS85FzjNDR}{uS85FzjNDR}), comparing four model variants: \textit{Llama-ZeroShot}, \textit{Llama-HighConf}, \textit{Llama-MultiRef}, and \textit{Llama-Consolidated}. As illustrated, \textit{Llama-Consolidated} produces more comprehensive feedback, synthesizing multifaceted critiques across strengths, weaknesses, and technical questions within a single coherent review.

In the example below, \textit{Llama-Consolidated} generates a more comprehensive review than the other variants, including specific details such as praising the ablation study while critiquing the lack of novelty and typographical errors. Conversely, \textit{Llama-MultiRef} lists only a single strength, and its proposed weaknesses lack deep substantiation. Similar shortcomings are observed in \textit{Llama-HighConf} and \textit{Llama-ZeroShot}, which generate only repetitive critiques.

\subsection{Image Captioning:}
\label{app:output-caption}
\begin{figure*}[htbp]
\centering
\begin{tcolorbox}[
    enhanced,
    boxrule=0.5pt,
    colframe=gray!40,      % Subtle gray border
    colback=gray!5,        % Very light gray background
    arc=2mm,
    boxsep=2mm,
    left=4mm, right=4mm, top=4mm, bottom=4mm,
    fontupper=\small\sffamily,
    width=0.9\textwidth    % Restrict width slightly for better readability
]
    
    % 1. Image Section
    \begin{center}
        \includegraphics[width=0.45\textwidth]{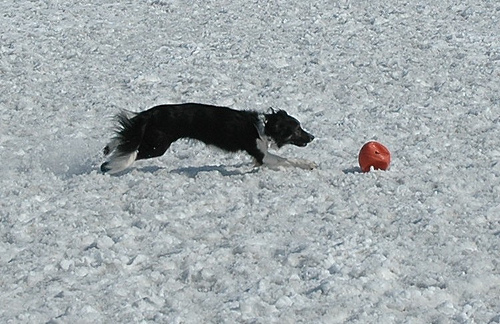}
    \end{center}
    
    \vspace{2mm}
    
    % 2. Human References Section
    \textbf{Human References:}
    \begin{enumerate}[label=(\arabic*), leftmargin=*, nosep, itemsep=2pt]
        \item A black and white dog is playing with an orange ball in the snow.
        \item The little black and white dog is in the snow with a red ball.
        \item A black-and-white dog goes for a flattened ball on the snow.
        \item A black dog is chasing a punctured ball in the snow.
        \item A dog getting a ball in the snow.
    \end{enumerate}
    
    \vspace{3mm}
    {\color{gray!50}\hrule} % Subtle divider line
    \vspace{3mm}
    
    % 3. Model Outputs Section
    \textbf{Model Generations:}
    \vspace{2mm}
    
    \textbf{PaliGemma-3B:} A dog running in the snow with a ball.
    
    \vspace{1.5mm}
    \textbf{Molmo-Multiref:} A dog chases after a red ball on the snow covered ground.
    
    \vspace{1.5mm}
    \textbf{Molmo-Consolidated:} A black and white dog chases a red ball in the snow.
    
    \vspace{1.5mm}
    \textbf{LLaVA-1.5-7B:} The image features a black and white dog playing with a red ball in a snowy field. The dog is actively chasing the ball, which is located towards the right side of the scene. The dog appears to be enjoying the game, as it runs across the snow-covered ground.
    
    \vspace{1.5mm}
    \textbf{ShareGPT4V-7B:} In the image, a black and white dog is the main subject, captured in the midst of a playful moment on a snowy beach. The dog is in motion, running towards the right side of the frame, its body angled towards a red ball that lies in the background. The ball, vibrant against the white snow, is positioned slightly to the right of the dog. The beach itself is blanketed in a layer of snow, with the dog's shadow stretching out behind it, adding depth to the scene. The image does not contain any text. The overall composition of the image suggests a lively and dynamic scene of a dog enjoying a playful moment in the snow.
    
\end{tcolorbox}
\vspace{-2mm}
\caption{Qualitative captioning example for image \texttt{3418504074.jpg}. The single unified box compares the five human references against outputs from short-form trained variants and off-the-shelf long-form captioners.}
\label{fig:flickr-qual-dog}
\end{figure*}

Figure~\ref{fig:flickr-qual-dog} shows a Karpathy-test image with the five human references and model captions.
Verbose systems (LLaVA, ShareGPT4V) produce much longer text than the short human references.

\section{Statistical Significance Tests}
\label{app:stats_results}
We conduct paired statistical significance testing across all evaluations to validate our findings. Specifically, the significance tests for the automatic evaluation for peer-review generation (from Tables~\ref{tab:raw_metrics} and \ref{tab:consolidated_metrics}) are detailed in Tables~\ref{tab:review_stats_original} and \ref{tab:review_stats_consolidated} while those for its ablation study are reported in Tables~\ref{tab:ablation_stats_original} and \ref{tab:ablation_stats_consolidated}. The significance tests for the image captioning task (from Tables~\ref{tab:flickr-human-refs} and ~\ref{tab:flickr-consolidated-refs}) are provided in Tables~\ref{tab:caption_stats_raw} and \ref{tab:caption_stats_consolidated}. 
\begin{table*}[htbp]
\centering
\resizebox{\textwidth}{!}{%
\begin{tabular}{l c ccc ccc ccc}
\toprule
 & & \multicolumn{3}{c}{\textbf{F1}} & \multicolumn{3}{c}{\textbf{Precision}} & \multicolumn{3}{c}{\textbf{Recall}} \\
\cmidrule(lr){3-5} \cmidrule(lr){6-8} \cmidrule(lr){9-11}
\textbf{Baseline} & \textbf{BS-F1} & \textbf{R-1} & \textbf{R-2} & \textbf{R-L} & \textbf{R-1} & \textbf{R-2} & \textbf{R-L} & \textbf{R-1} & \textbf{R-2} & \textbf{R-L} \\
\midrule
Llama-ZeroShot & +2.17** & +12.05** & +6.06** & +3.42** & -3.81** & +1.01** & -4.50** & +17.45** & +7.65** & +6.22** \\
Llama-HighConf & +2.36** & +15.70** & +7.99** & +5.27** & +1.20** & +3.55** & -2.34** & +18.69** & +8.73** & +6.90** \\
Reviewer2 & +3.14** & +33.07** & +13.33** & +11.44** & -6.63** & +0.51** & -13.19** & +36.32** & +14.09** & +13.99** \\
OpenReviewer & +2.07** & +14.40** & +7.34** & +4.83** & -2.03** & +2.33** & -3.52** & +18.87** & +8.53** & +7.19** \\
Llama-MultiRef & +2.31** & +16.71** & +8.37** & +5.63** & -0.51** & +3.12** & -3.44** & +20.61** & +9.35** & +7.79** \\
SEA-E & +1.05** & +3.21** & +5.36** & +3.76** & +4.18** & +6.25** & +4.45** & +2.32** & +4.66** & +3.18** \\
AutoRev & +1.25** & +2.77** & +1.12** & +0.26** & -2.09** & -0.63** & -2.08** & +5.34** & +2.04** & +1.51** \\
\bottomrule
\end{tabular}
}
\caption{Paired significance tests of Llama-Consolidated against each baseline on the consolidated reviews. Entries are score differences (consolidated - baseline). significant differences ($p < 0.01$) are denoted with **.}
\label{tab:review_stats_consolidated}
\end{table*}

\begin{table*}[htbp]
\centering
\resizebox{\textwidth}{!}{%
\begin{tabular}{l c ccc ccc ccc}
\toprule
 & & \multicolumn{3}{c}{\textbf{F1}} & \multicolumn{3}{c}{\textbf{Precision}} & \multicolumn{3}{c}{\textbf{Recall}} \\
\cmidrule(lr){3-5} \cmidrule(lr){6-8} \cmidrule(lr){9-11}
\textbf{Baseline} & \textbf{BS-F1} & \textbf{R-1} & \textbf{R-2} & \textbf{R-L} & \textbf{R-1} & \textbf{R-2} & \textbf{R-L} & \textbf{R-1} & \textbf{R-2} & \textbf{R-L} \\
\midrule
Llama-ZeroShot & +0.53** & -0.78** & -0.04 & -1.25** & -9.95** & -2.44** & -5.59** & +9.63** & +2.74** & +3.82** \\
Llama-HighConf & +0.26** & +1.25** & +0.47** & -0.09 & -8.44** & -2.15** & -4.89** & +10.12** & +2.91** & +4.25** \\
Reviewer2 & +0.46** & +10.65** & +2.85** & +2.81** & -21.62** & -5.98** & -15.52** & +27.98** & +7.51** & +11.98** \\
OpenReviewer & +0.06** & -0.54** & -0.18** & -0.74** & -10.96** & -2.96** & -5.81** & +9.74** & +2.58** & +4.36** \\
Llama-MultiRef & +0.11** & +0.70** & +0.26** & -0.30** & -10.83** & -2.86** & -6.04** & +11.17** & +3.12** & +4.95** \\
SEA-E & +0.22** & -0.93** & +1.13** & +1.54** & -0.60** & +1.07** & +1.46** & -1.79** & +1.29** & +1.69** \\
AutoRev & +0.53** & +0.17** & +0.31** & -0.42** & -2.14** & -0.30** & -1.46** & +3.90** & +1.33** & +1.30** \\
\bottomrule
\end{tabular}
}
\caption{Paired significance tests of Llama-Consolidated against each baseline on the original unconsolidated reviews. Entries are score differences (consolidated - baseline). significant differences ($p < 0.01$) are denoted with **}
\label{tab:review_stats_original}
\end{table*}

\begin{table*}[htbp]
\centering
\resizebox{\textwidth}{!}{%
\begin{tabular}{l c ccc ccc ccc}
\toprule
 & & \multicolumn{3}{c}{\textbf{F1}} & \multicolumn{3}{c}{\textbf{Precision}} & \multicolumn{3}{c}{\textbf{Recall}} \\
\cmidrule(lr){3-5} \cmidrule(lr){6-8} \cmidrule(lr){9-11}
\textbf{Baseline (MultiRef)} & \textbf{BS-F1} & \textbf{R-1} & \textbf{R-2} & \textbf{R-L} & \textbf{R-1} & \textbf{R-2} & \textbf{R-L} & \textbf{R-1} & \textbf{R-2} & \textbf{R-L} \\
\midrule
Llama-3.1-8B & +2.58** & +16.42** & +5.36** & +4.39** & +0.44** & +1.96** & -3.39** & +20.47** & +6.10** & +6.44** \\
Mistral-7B-v0.3 & +1.92** & +12.74** & +5.58** & +3.91** & +2.87** & +3.22** & -0.96** & +14.46** & +5.89** & +4.98** \\
Qwen-2.5-7B & +2.15** & +13.59** & +4.38** & +4.14** & +2.51** & +2.34** & -1.39** & +15.40** & +4.56** & +5.21** \\
\bottomrule
\end{tabular}
}
\caption{Paired significance tests of each consolidated model against its MultiRef counterpart on the consolidated reviews. significant differences ($p < 0.01$) are denoted with **}
\label{tab:ablation_stats_consolidated}
\end{table*}

\begin{table*}[htbp]
\centering
\resizebox{\textwidth}{!}{%
\begin{tabular}{l c ccc ccc ccc}
\toprule
 & & \multicolumn{3}{c}{\textbf{F1}} & \multicolumn{3}{c}{\textbf{Precision}} & \multicolumn{3}{c}{\textbf{Recall}} \\
\cmidrule(lr){3-5} \cmidrule(lr){6-8} \cmidrule(lr){9-11}
\textbf{Baseline (MultiRef)} & \textbf{BS-F1} & \textbf{R-1} & \textbf{R-2} & \textbf{R-L} & \textbf{R-1} & \textbf{R-2} & \textbf{R-L} & \textbf{R-1} & \textbf{R-2} & \textbf{R-L} \\
\midrule
Llama-3.1-8B & +0.70** & +5.57** & +1.10** & +0.58** & -6.31** & -1.23** & -4.70** & +17.88** & +3.50** & +5.81** \\
Mistral-7B-v0.3 & +0.29** & +3.53** & +0.73** & +0.56** & -3.75** & -0.88** & -2.74** & +10.44** & +2.29** & +3.70** \\
Qwen-2.5-7B & +0.62** & +6.15** & +1.17** & +1.16** & -2.78** & -0.45** & -2.91** & +14.49** & +2.69** & +4.77** \\
\bottomrule
\end{tabular}
}
\caption{Paired significance tests of each consolidated model against its MultiRef counterpart on the original unconsolidated reviews. Significant differences ($p < 0.01$) are denoted with **.}
\label{tab:ablation_stats_original}
\end{table*}

\begin{table*}[htbp]
\centering
\resizebox{\textwidth}{!}{%
\begin{tabular}{l c ccc ccc ccc}
\toprule
 & & \multicolumn{3}{c}{\textbf{F1}} & \multicolumn{3}{c}{\textbf{Precision}} & \multicolumn{3}{c}{\textbf{Recall}} \\
\cmidrule(lr){3-5} \cmidrule(lr){6-8} \cmidrule(lr){9-11}
\textbf{Baseline} & \textbf{LCS} & \textbf{R-1} & \textbf{R-2} & \textbf{R-L} & \textbf{R-1} & \textbf{R-2} & \textbf{R-L} & \textbf{R-1} & \textbf{R-2} & \textbf{R-L} \\
\midrule
BLIP-2-6.7B & +0.02** & +0.07** & +0.06** & +0.04** & -0.10** & -0.00 & -0.10** & +0.16** & +0.09** & +0.12** \\
GIT-Base & +0.05** & +0.14** & +0.09** & +0.10** & -0.08** & +0.02** & -0.09** & +0.23** & +0.13** & +0.18** \\
LLaVA-1.5-7B & +0.07** & +0.23** & +0.13** & +0.21** & +0.35** & +0.17** & +0.30** & -0.09** & +0.02** & -0.03** \\
MultiRef & +0.02** & +0.07** & +0.04** & +0.04** & -0.13** & -0.04** & -0.14** & +0.16** & +0.08** & +0.12** \\
PaliGemma-3B & +0.04** & +0.08** & +0.06** & +0.05** & -0.15** & -0.04** & -0.16** & +0.19** & +0.10** & +0.14** \\
ShareGPT4V-7B & +0.06** & +0.30** & +0.15** & +0.26** & +0.41** & +0.19** & +0.34** & -0.13** & +0.00 & -0.09** \\
Molmo-SingleRef & +0.03** & +0.08** & +0.05** & +0.04** & -0.12** & -0.04** & -0.14** & +0.17** & +0.09** & +0.13** \\
Molmo-ZeroShot & +0.07** & +0.24** & +0.14** & +0.23** & +0.37** & +0.18** & +0.32** & -0.17** & +0.01** & -0.07** \\
\bottomrule
\end{tabular}
}
\caption{Paired significance tests of Molmo-Consolidated against each baseline on the consolidated captions. Entries are score differences (Consolidated - baseline). Significant differences ($p < 0.01$) are denoted with **.}
\label{tab:caption_stats_consolidated}
\end{table*}

\begin{table*}[htbp]
\centering
\resizebox{\textwidth}{!}{%
\begin{tabular}{l c ccc ccc ccc}
\toprule
 & & \multicolumn{3}{c}{\textbf{F1}} & \multicolumn{3}{c}{\textbf{Precision}} & \multicolumn{3}{c}{\textbf{Recall}} \\
\cmidrule(lr){3-5} \cmidrule(lr){6-8} \cmidrule(lr){9-11}
\textbf{Baseline} & \textbf{LCS} & \textbf{R-1} & \textbf{R-2} & \textbf{R-L} & \textbf{R-1} & \textbf{R-2} & \textbf{R-L} & \textbf{R-1} & \textbf{R-2} & \textbf{R-L} \\
\midrule
BLIP-2 (6.7B) & +0.02** & -0.03** & -0.01** & -0.03** & -0.11** & -0.04** & -0.10** & +0.12** & +0.04** & +0.09** \\
GIT-Base & +0.05** & +0.02** & +0.01** & -0.00 & -0.12** & -0.04** & -0.12** & +0.21** & +0.08** & +0.16** \\
LLaVA-1.5 (7B) & +0.07** & +0.22** & +0.09** & +0.19** & +0.22** & +0.09** & +0.19** & -0.08** & -0.00 & -0.05** \\
MultiRef & +0.02** & -0.04** & -0.03** & -0.05** & -0.14** & -0.07** & -0.14** & +0.12** & +0.03** & +0.09** \\
PaliGemma (3B) & +0.04** & -0.04** & -0.02** & -0.05** & -0.16** & -0.07** & -0.16** & +0.15** & +0.05** & +0.11** \\
ShareGPT4V (7B) & +0.06** & +0.27** & +0.11** & +0.24** & +0.25** & +0.10** & +0.22** & -0.09** & -0.00 & -0.09** \\
Molmo-SingleRef & +0.03** & -0.03** & -0.02** & -0.04** & -0.14** & -0.06** & -0.14** & +0.14** & +0.04** & +0.10** \\
Molmo-ZeroShot & +0.07** & +0.24** & +0.10** & +0.21** & +0.23** & +0.09** & +0.20** & -0.13** & +0.02** & -0.07** \\
\bottomrule
\end{tabular}
}
\caption{Paired significance tests of Molmo-Consolidated against each baseline on the original human written captions. Entries are score differences (Consolidated - baseline). Significant differences ($p < 0.01$) are denoted with **.}
\label{tab:caption_stats_raw}
\end{table*}
\section{Human Evaluation: Supplementary Details}
\label{app:human_eval}
\subsection{Annotation Guidelines and Procedure}
\begin{tcolorbox}[title={Guidelines (Peer Review Generation Task)}, width=\linewidth, colback=white, colframe=gray, arc=0pt, outer arc=5pt, boxrule=0.5pt, leftrule=2pt, rightrule=2pt, right=4pt, left=4pt, top=4pt, bottom=4pt, toprule=0pt, bottomrule=2pt]
\label{box:annotation_instructions}
You have 5 papers to evaluate. For each paper, you will go through 4 sections in order:

\begin{quote}
Summary $\rightarrow$ Strengths $\rightarrow$ Weaknesses $\rightarrow$ Questions
\end{quote}

For each section:

\begin{enumerate}
    \item Read all the original reviewer comments shown at the top. These are the original reviews submitted by human reviewers for that paper.
    \item Compare the 4 model outputs shown as tabs (Model A, B, C, D). These are system-generated reviews.
    \item Assign a rank to each model using the dropdowns:
    \begin{itemize}
        \item \texttt{1} = best, \texttt{4} = worst
        \item Ties are allowed --- give two models the same rank if you think they are equally good (e.g. both get \texttt{1}, and the next best gets \texttt{2})
        \item Note: The best reviews are those which cover most of the points from the original reviewer comments (Comprehensiveness)
    \end{itemize}
\end{enumerate}

Repeat for all 4 sections, then move to the next paper.
\end{tcolorbox}
For peer-review generation, all 30 annotators are graduate researchers familiar with the machine learning peer-review process. The evaluation interface presents four anonymized model outputs side-by-side for each review section of a given paper, with model identities masked behind randomly assigned labels. Annotators are instructed to rank the four outputs from 1 (best) to 4 (worst) based on \textbf{Coverage (Comprehensiveness)}. This measures the degree to which the generated section captures the breadth of critiques, strengths, or questions diverse aspects of the original peer reviews. Each annotator evaluates 5 papers across 4 sections (20 judgements per annotator). 

\begin{tcolorbox}[title={Guidelines (Image Captioning Task)}, width=\linewidth, colback=white, colframe=gray, arc=0pt, outer arc=5pt, boxrule=0.5pt, leftrule=2pt, rightrule=2pt, right=4pt, left=4pt, top=4pt, bottom=4pt, toprule=0pt, bottomrule=2pt]
\label{box:annotation_instructions_captioning}
For \textbf{each image}, please follow these steps:

\begin{enumerate}
    \item Look at the source image provided on the left.
    \item Compare the \textbf{4 anonymized captions} displayed on the right.
    \item \textbf{Assign a rank} to each model caption from \textbf{1 (Best)} to \textbf{4 (Worst)}.
    \begin{itemize}
        \item \textbf{Ties are allowed} : use the same number to indicate a tie (e.g., both can be ranked \texttt{1}).
    \end{itemize}
\end{enumerate}

\vspace{1em}
\textbf{Guideline for Correctness (Holistic Quality):}\\
Evaluate how well the caption accurately and naturally describes the most salient features of the image, without hallucinating non-existent details. Rank the captions based on their overall correctness and quality.
\end{tcolorbox}

For image captioning, annotators validate the generated captions based on their correctness, which encapsulates how accurately and naturally the text describes the most salient visual features without hallucinating non-existent details. The evaluation interface presents four anonymized models, with their true identities masked behind randomly assigned labels. Annotators are instructed to rank the four outputs from 1 (best) to 4 (worst), with fractional ties allowed to accommodate captions of comparable quality.

All annotators provide informed consent ensuring their data is used exclusively for this study. All collected data is fully anonymized, and participants receive fair compensation for completing the evaluation tasks.

\subsection{Annotator Preference and Ranking}

\begin{figure}[htbp]
    \centering
    \includegraphics[width=\linewidth]{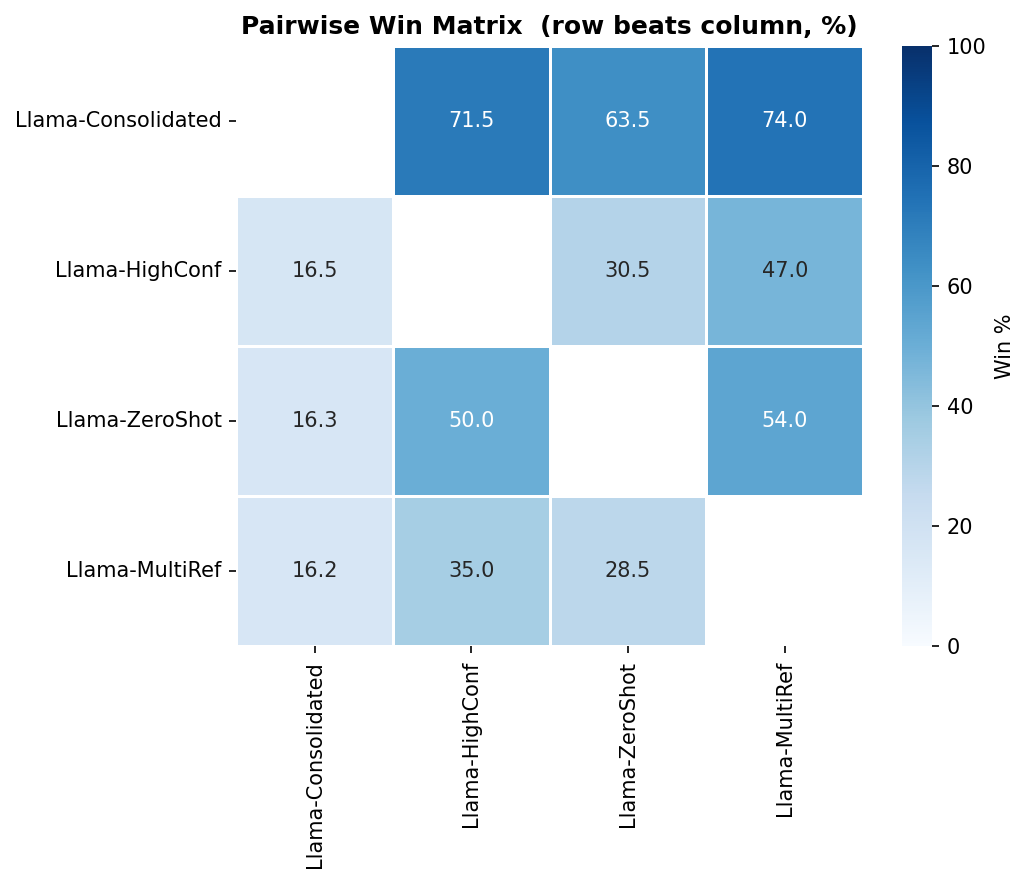}
    \caption{Pairwise win matrix indicating the percentage of instances in which the row model achieved a higher rank than the column model. \textit{Llama-Consolidated} is ranked above the baselines in more head-to-head comparisons.}
    \label{fig:win_matrix}
\end{figure}

\begin{figure}[htbp]
    \centering
    \includegraphics[width=\linewidth]{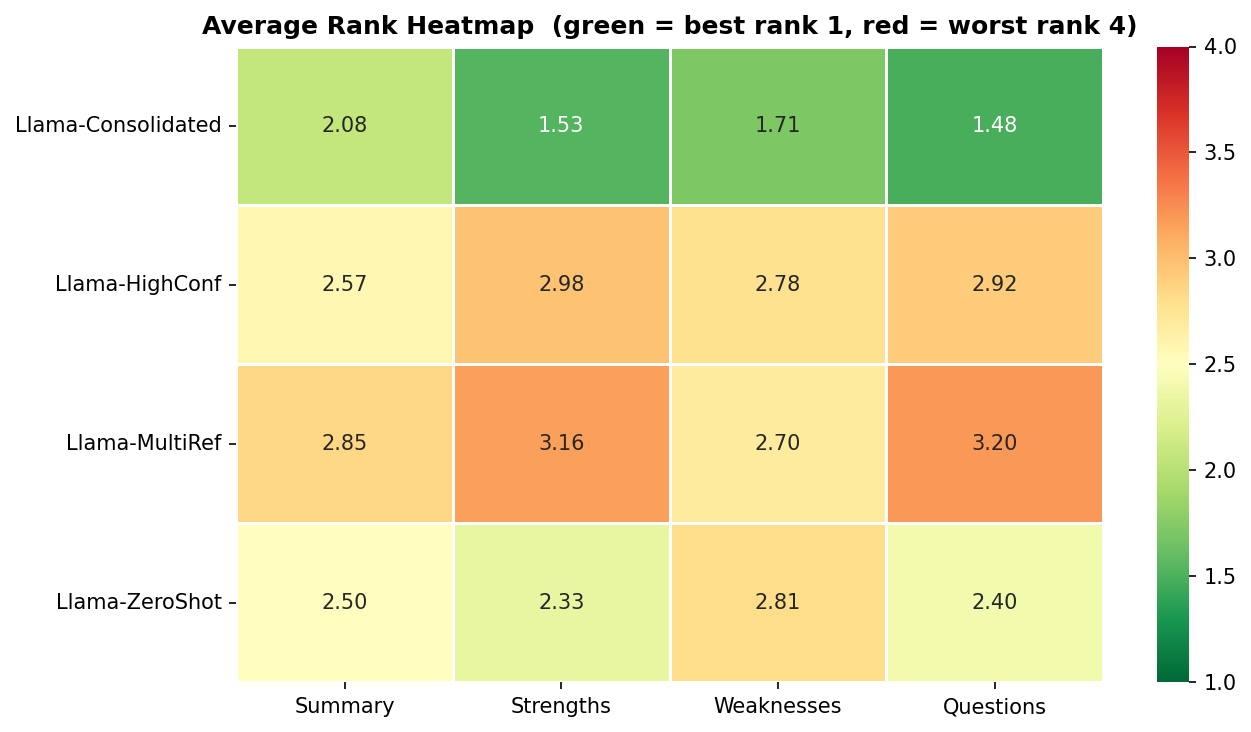}
    \caption{Heatmap of the average model rank across different review sections. Lower average ranks (indicating better performance) are highlighted in green, while higher ranks are shown in red.}
    \label{fig:rank_heatmap}
\end{figure}

\begin{figure}[htbp]
    \centering
    \includegraphics[width=\linewidth]{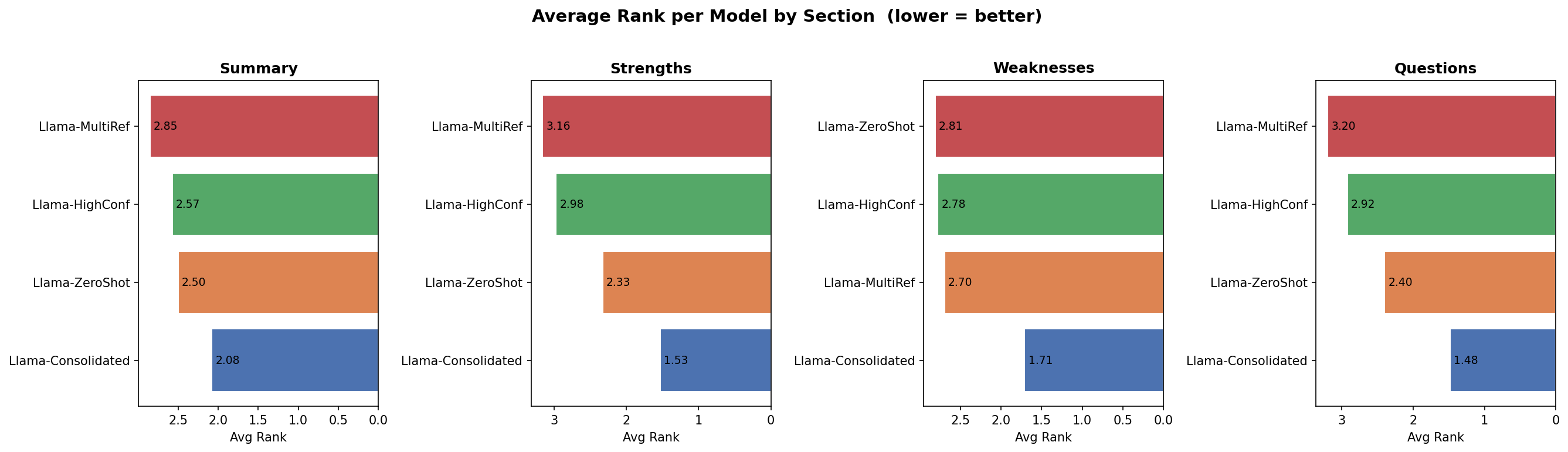}
    \caption{Average annotator rank per review section where lower rank is better. \textit{Llama-Consolidated} ranks first in all four sections, with the largest margin in \textit{Strengths} and \textit{Questions}.}
    \label{fig:avg_rank}
\end{figure}

For the peer-review generation task, as discussed in Section~\ref{main:human_eval_result} and illustrated in Figure~\ref{fig:avg_rank}, \textit{Llama-Consolidated} achieves the best overall average rank of 1.70 (where lower is better). It is followed by the zero-shot baseline, \textit{Llama-ZeroShot} (2.51), and the single-reference model, \textit{Llama-HighConf} (2.81). The unconsolidated multi-reference approach, \textit{Llama-MultiRef}, performs the worst overall with an average rank of 2.98. The pairwise win matrix in Figure~\ref{fig:win_matrix} reinforces this advantage, demonstrating that \textit{Llama-Consolidated} wins most head-to-head comparisons against all other system types. This shows that synthesizing diverse reviewer perspectives into a single, unified target provides a substantially cleaner and more comprehensive reference than either forcing a single viewpoint or exposing the model to the contradictory noise of multiple original reviews. Furthermore, Figure~\ref{fig:rank_heatmap} visualizes the average rank per model across individual sections. This shows that the empirical advantage of consolidated training is robust across all facets of the peer review process, from summarization to highly specific critiques and questions.

\begin{figure}[htbp]
    \centering
    \includegraphics[width=\linewidth]{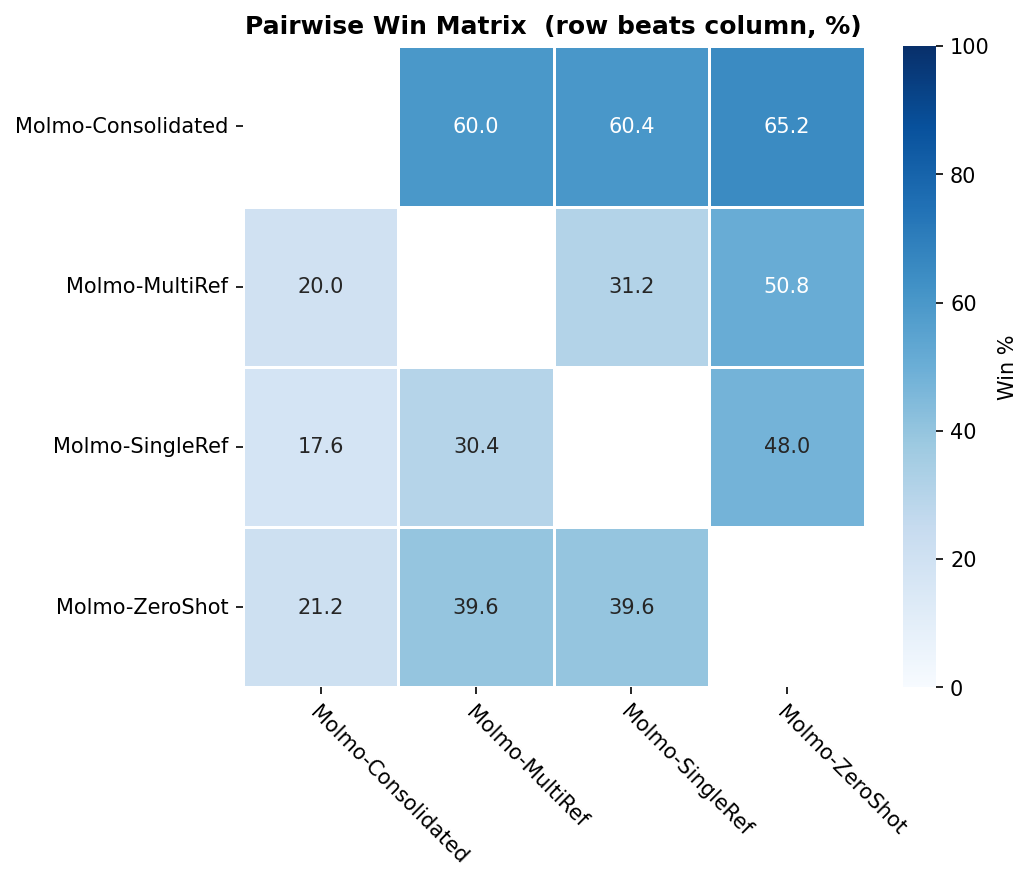}
    \caption{Pairwise win matrix indicating the percentage of instances in which the row model achieved a higher rank than the column model. \textit{Molmo-Consolidated} is ranked above the baselines in more head-to-head comparisons.}
    \label{fig:win_matrix_image}
\end{figure}

Similarly for the image-captioning task, Figure~\ref{fig:win_matrix_image} also shows \textit{Molmo-Consolidated} paradigm is ranked above all baselines in 60.0-65.2\% of comparisons. In contrast, the alternative paradigms fail to establish a clear ordering between one another; for instance, \textit{Multi-Ref} beats \textit{Single-Ref} in only 31.2\% of cases, while \textit{Single-Ref} wins 30.4\% of the time, indicating a high tie rate and comparable generation quality between the two. This reinforces our claim that consolidating diverse references yields outputs that are preferred by humans.

\subsection{Inter-Annotator Agreement}
% \begin{table}[htbp]
% \centering
% \small
% \begin{tabular}{lcc}
% \toprule
% \textbf{Section} & \textbf{Kendall's $W$} & \textbf{Krippendorff's $\alpha$} \\
% \midrule
% Summary    & 0.619 & 0.363 \\
% Strengths  & \textbf{0.690} & \textbf{0.487} \\
% Weaknesses & 0.577 & 0.301 \\
% Questions  & 0.662 & 0.440 \\
% \midrule
% \textbf{Overall} & \textbf{0.637} & 0.398 \\
% \bottomrule
% \end{tabular}
% \caption{Inter-annotator agreement by section. The Weaknesses section shows the lowest agreement across both metrics, consistent with the inherent subjectivity of identifying missed contributions in a scientific paper.}
% \label{tab:iaa_full}
% \end{table}
\begin{figure}[htbp]
    \centering
    \includegraphics[width=\linewidth]{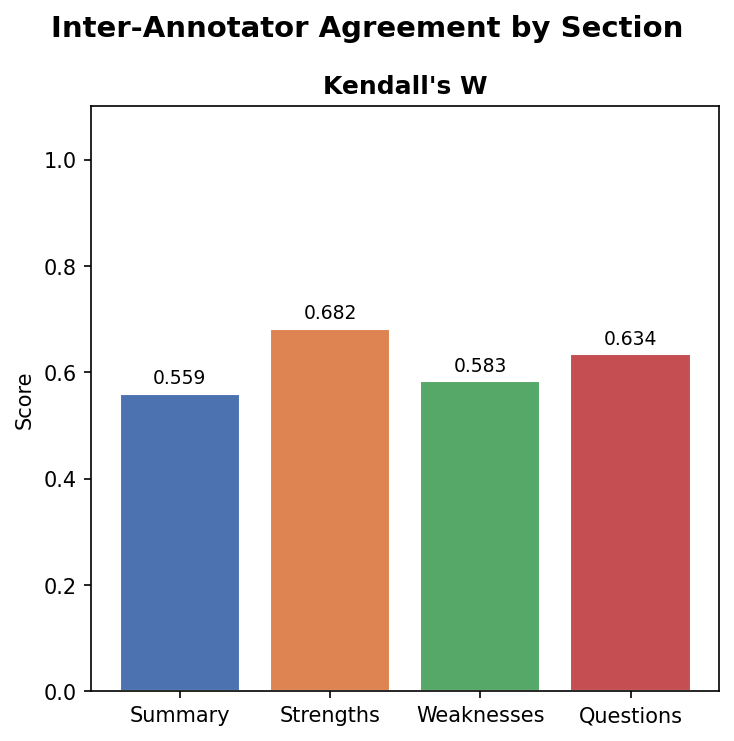}
    \caption{Inter-annotator agreement denoted by Kendall's $W$ per review section.}
    \label{fig:iaa_section}
\end{figure}
For peer-review generation, Figure~\ref{fig:iaa_section} provides a visual breakdown of inter-annotator agreement for each section of the review. Regarding model performance across these sections, \textit{Llama-Consolidated} demonstrates its strongest results in the highly specific critique categories, achieving average ranks of 1.53 for \textit{Strengths} and 1.48 for \textit{Questions}. In contrast, \textit{Llama-MultiRef} struggles in these exact areas, scoring 3.16 and 3.20, respectively. This indicates that nuanced, specific critiques are far better captured by \textit{Llama-Consolidated} compared to the baseline systems.

% In the Weaknesses section, \textit{Llama-Consolidated} remains the top performer (1.71). Interestingly, \textit{Llama-MultiRef} performs slightly better here (2.70) than the other baselines, narrowly beating \textit{Llama-HighConf} (2.78) and \textit{Llama-Inference} (2.81).
The performance margin between models is narrowest in the \textit{Summary} section. While \textit{Llama-Consolidated} still leads (2.08), \textit{Llama-ZeroShot} (2.50) and \textit{Llama-HighConf} (2.57) perform competitively. This narrower gap is expected, as summaries inherently exhibit less variance and require less complex reasoning compared to the other critical evaluation sections of a peer review.

\subsection{Conference-Level Breakdown}
\begin{figure*}[htbp]
    \centering
    \includegraphics[width=0.8\linewidth]{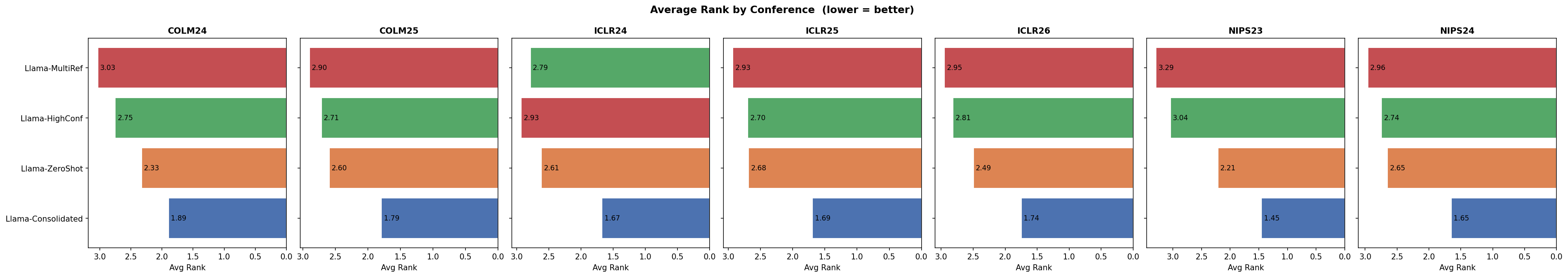}
    \caption{Average ranks stratified by conference (ICLR 2024, ICLR 2025, ICLR 2026, NeurIPS 2023, NeurIPS 2024, COLM 2024, COLM 2025).}
    \label{fig:conference}
\end{figure*}
Figure~\ref{fig:conference} demonstrates that \textit{Llama-Consolidated} maintains the top rank across all seven venues. While the rankings of the remaining three models fluctuate slightly depending on the venue, \textit{Llama-MultiRef} remains the worst-performing model in the majority of conferences.

\newpage

\end{document}